\documentclass[runningheads]{llncs}
\usepackage{graphicx}
\usepackage{amsmath}
\usepackage{hyperref}
\usepackage{appendix}
\usepackage{multirow}
\usepackage{tabularx}
\usepackage{makecell}
\usepackage{algorithm}
\usepackage{algorithmic}
\usepackage{xcolor}
\usepackage{marvosym}
\hypersetup{
    colorlinks=true,
    linkcolor=red,
    urlcolor=blue,
    citecolor=green
}

\begin{document}

\title{
    FormalGeo: An Extensible Formalized Framework For Olympiad Geometric Problem Solving 
}
\titlerunning{FormalGeo}

\author{
    Xiaokai Zhang\inst{1} \and
    Na Zhu\inst{1,2} \and
    Yiming He\inst{1,2} \and
    Jia Zou\inst{1,2} \and
    Qike Huang\inst{1,2} \and
    Xiaoxiao Jin\inst{1,2} \and
    Yanjun Guo\inst{1,2} \and
    Chenyang Mao\inst{1,2} \and
    Yang Li\inst{1} \and
    Zhe Zhu\inst{1,2} \and
    Dengfeng Yue\inst{1,2} \and
    Fangzhen Zhu\inst{1} \and
    Yifan Wang\inst{1,2} \and
    Yiwen Huang\inst{1} \and
    Runan Wang\inst{1,2} \and
    Cheng Qin\inst{1,2} \and
    Zhenbing Zeng\inst{3} \and
    Shaorong Xie\inst{1} \and
    Xiangfeng Luo\inst{1} \and
    Tuo Leng\inst{1,2}\textsuperscript{(\Letter)}
}
\authorrunning{X. Zhang et al.}

\institute{
    School of Computer Engineering and Science, Shanghai University, Shanghai, China \\ \email{tleng@shu.edu.cn} \and
    Institute of Artificial Intelligence, Shanghai University, Shanghai, China \and
    College of Sciences, Shanghai University, Shanghai, China
}

\maketitle

\begin{abstract}
This is the first paper in a series of work we have accomplished over the past three years. In this paper, we have constructed a consistent plane geometry formal system. This will serve as a crucial bridge between IMO-level plane geometry challenges and readable AI automated reasoning. Within this formal framework, we have been able to seamlessly integrate modern AI models with our formal system. AI is now capable of providing deductive reasoning solutions to IMO-level plane geometry problems, just like handling other natural languages, and these proofs are readable, traceable, and verifiable.
We propose the \emph{geometry formalization theory} (GFT) to guide the development of the geometry formal system. 
Based on the GFT, we have established the \emph{FormalGeo}, which consists of 88 geometric predicates and 196 theorems. It can represent, validate, and solve IMO-level geometry problems.
we also have crafted the \emph{FGPS} (formal geometry problem solver) in Python. It serves as both an interactive assistant for verifying problem-solving processes and an automated problem solver.
We've annotated the \emph{formalgeo7k} and \emph{formalgeo-imo} datasets. The former contains 6,981 (expand to 133,818 through data augmentation) geometry problems, while the latter includes 18 (expand to 2,627 and continuously increasing) IMO-level challenging geometry problems. All annotated problems include detailed formal language descriptions and solutions.
\emph{AI} can be integrated into formal systems in various roles. It can act as a parser, enabling the autoformalization of natural language and geometric diagrams. It can also serve as a solver, running search tree pruning.
Implementation of the formal system and experiments validate the correctness and utility of the GFT. The backward depth-first search method only yields 2.42\% problem-solving failure rate on formalgeo7k. We can incorporate deep learning techniques to achieve lower one. The source code of FGPS and datasets are available \href{https://github.com/BitSecret/FGPS}{here}.


\end{abstract}

\section{Introduction}

Since the inception of mathematics, it has inherently encompassed both structure and computation. These two facets not only interact with each other but also mutually reinforce each other. With the advent of modern computing, it has started to exert its influence on mathematics in two distinct ways. On one hand, it serves as a mathematical tool, empowering mathematical computations more than ever before, thereby directly impacting the balance of values and methodologies within mathematics. On the other hand, as a mathematical medium, it indirectly reshapes the content and structure of mathematics through innovative applications, ushering in an unprecedented era of prosperity and development for the field.

Within this transformative landscape, the field of mathematical mechanization emerged, situated at the intersection of mathematics and computer science. The translation and conversion of mathematical knowledge into a language comprehensible to computers are evidently the first and indispensable steps in the fusion of mathematics and computer science. This is the essence of formal mathematics.

Formal mathematics serves as the foundation for computer-aided mathematical problem-solving. It employs a symbol system that adheres to a particular artificial grammar, enabling the representation of any concept, proposition, and inference. It is only when mathematical knowledge is rigorously formalized that the problem-solving process in mathematics can be described as a deterministic algorithm and implemented within a computer. Over the course of several decades, numerous formal mathematics systems and tools have emerged, such as Lean, Isabelle and Coq.

The development of artificial intelligence (AI) has introduced a new paradigm for computer-aided mathematical problem-solving. AI-assisted mathematical problem solving is a rapidly developing area which aims to apply deep learning technology to math problem. AI systems can assume various roles within formal mathematical systems. They can serve as mathematical problem solvers \cite{zheng2022miniff,polu2022formal,jiang2022thor,lample2022hypertree,jiang2022draft}, automatically generating problem solutions. They can act as mathematical problem parsers \cite{wu2022autoformalization,cunningham2022towards,zhang2022plane}, assisting humans in converting mathematical knowledge into formal descriptions. Moreover, they can even take on the role of mathematical problem proposers \cite{davies2021advancing,raayoni2021generating,mishra2023mathematical}, suggesting mathematical conjectures.

A consistent mathematical formal system is an indispensable component of AI-assisted mathematical problem-solving. On the one hand, a unified format for mathematical problem descriptions and datasets helps in avoiding interference from other factors and serves as a standard for evaluating the capabilities of AI models. On the other hand, the answers generated by AI models must be verified for credibility, and manual verification is both inefficient and error-prone. This necessitates the capability for computers to automatically validate the answers. The Stanford 2021 AI100 report \cite{littman2022gathering} designates the IMO grand challenge \cite{daniel2019imo} as a landmark event in the development of artificial intelligence: To create an AI system capable of receiving a formal problem representation, generating a formal (i.e., machine-checkable) proof for the problem, and attaining a gold medal in the International Mathematical Olympiad. AIMO Prize \cite{xtx2023artificial} was established to motivate the development of AI models that are capable of winning a gold medal in the IMO. Serving as a bridge that connects the two crucial research domains of AI and mathematics, formal mathematics has garnered increasing attention from researchers.

\begin{figure}
\includegraphics[width=\textwidth]{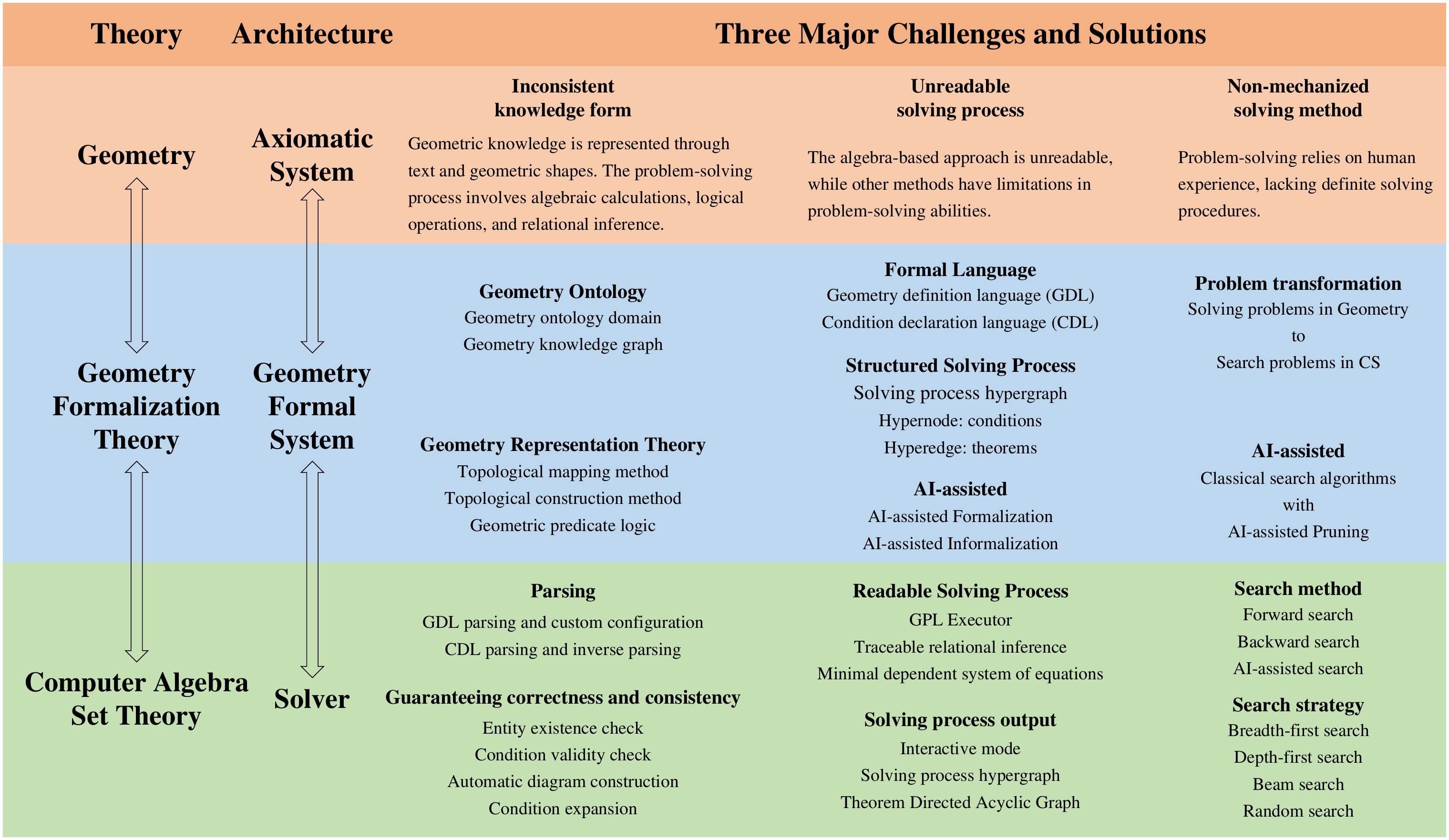}
\caption{Three major challenges and solutions in formal plane geometry.}
\label{fig-three_major_challenges}
\end{figure}

The intersection and fusion of AI and formal mathematics have yielded a series of achievements. However, as an essential branch of formal mathematics, the formalization and mechanized solving of geometric problems have still made slow progress. As depicted in Fig.~\ref{fig-three_major_challenges}, this domain has long been constrained by three major challenges: inconsistent knowledge form, unreadable
solving process and non-mechanized solving method.

Addressing the first challenge, we introduced the geometry formalization theory (GFT) to unify the representation of geometric knowledge, including geometry ontology and geometry representation theory. geometry ontology provides a comprehensive overview of plane geometry from a highly abstract philosophical perspective, creating the geometry ontology domain. The geometry ontology domain consists of four quadrants based on the dimensions of number-shape and dynamic-static. Each quadrant maps the relationships between modern geometric axiom systems, geometry formal systems, and solvers across three hierarchical levels. Leveraging the geometry ontology domain, we can ensure the comprehensiveness of formal system design. Geometry representation theory investigates how to represent static geometric knowledge and dynamic problem-solving processes. Based on topological mapping method and topological construction method, we can transform geometric diagrams into textual or symbolic descriptions. Using geometry predicate logic, we can unify relation reasoning, logical operations, and algebraic calculations into rigorous formal representations.

Addressing the second challenge, we designed a set of geometry formal languages to serve as a bridge for communication between humans and computers. geometry formal languages consist of geometry definition language and condition declaration language, where the former is used to personalize the configuration of solvers and import into the formal system, and the latter is used for inputting problem descriptions. Geometry formal languages possess rigorous syntactic descriptions that can be mechanically processed by computers, while their syntax is similar to predicate logic, providing good readability. We abstract the process of geometry problem-solving as a hyper tree (as shown in Fig.~\ref{fig-problem_example}), where tree nodes represent known conditions, tree edges represent geometric theorems, and the problem-solving process is a path from the root node to the target leaf node. We can also utilize modern AI techniques to automatically translate between natural language and formal language, further enhancing readability.

Addressing the third challenge, we transformed the mathematical problems of geometric problem-solving into computational search problems in the field of computer science, achieving both forward search and backward search. Forward search starts with known conditions and continuously applies theorems to obtain new conditions until the target condition is achieved. Backward search starts from the problem goal, expands the goal into multiple subgoals according to theorems, and repeats this process until all new subgoals are known conditions. The vast search space gives rise to the problem of combinatorial explosion, with the search time for difficult problems exhibiting exponential growth. Pruning of the search tree is essential. In addition to classical pruning techniques such as Monte Carlo Tree Search and Alpha-Beta pruning, we leverage deep learning as a powerful pruning method to accelerate the solving process.

Our contributions can be summarized as follows:

1.We propose the GFT, which comprises geometry ontology and geometry representation theory. This theory provides a comprehensive framework for the field of plane geometry, unifying the representation of symbolic geometric knowledge and graphic geometric knowledge. It encompasses various operations, including relation reasoning, logical operations, and algebraic calculations, within a unified representation framework.

2.We have established the \emph{FormalGeo} based on the GFT, which consists of 88 geometric predicates and 196 theorems. It can represent, validate, and solve geometry problems from STA-level to IMO-level.

3.We have crafted the formal geometry problem solver (FGPS) in Python. It serves as both an interactive assistant for verifying problem-solving processes and an automated problem solver, utilizing various methods such as forward search, backward search and AI-assisted search and various strategy such as depth-first search, breadth-first search, random search and beam search. FGPS incorporates features such as formal statement parsing, condition validity checks, automatic diagram construction, and condition expansion. It is capable of executing backtrackable, interpretable algebraic equation solving and relational reasoning.

4.We've annotated the formalgeo7k and formalgeo-imo datasets. The former contains 6,981 (expand to 133,818) geometry problems, while the latter includes 18 (expand to 2,627 and continuously increasing) IMO-level challenging geometry problems. Each problem comprises a complete natural language description, geometric shapes, formal language annotations, and theorem sequences annotations.

5.We conducted experiments on the formalgeo7k, comparing two search-based problem-solving methods (forward and backward) and 4 search strategies (breadth-first, depth-first, random, beam) in terms of their success rate, time-consuming, and search step. The forward random search method yields a 39.7\% problem-solving accuracy rate, and we can incorporate deep learning techniques to achieve higher one. 

\section{Related Work}

Gelernter et al. developed the pioneering automated geometry problem-solving system known as the Geometry Theorem Prover \cite{gelernter1959realization}, which employed a backward search approach to solve pre-formalized problems. Nevins pointed out that the forward chaining method \cite{nevins1975plane} can also be effective by efficiently representing the known conditions of the problem and limiting the typical application of those conditions. The development of geometry problem solving has led to the emergence of various downstream tasks, including geometry problem formalization \cite{gan2018automatic,rao2022method},geometric knowledge extraction \cite{sachan2017textbooks,sachan2020discourse,yu2021geore,huang2022novel,zhou2022research}, geometric diagram parsing \cite{seo2014diagram,zhang2022plane,wong2022euclidnet}, geometric theorem proving \cite{yu2019framework,gan2019automatically,kovacs2022automated}, and geometry problem solving \cite{seo2015solving,zhong2015interactive,alvin2014synthesis,alvin2017synthesis,sachan2017learning,yu2017understanding}.

Wen-Tsun proposed the Wu's Method \cite{wu1978decision}, which transforms geometry problem into a system of algebraic equations consisting of polynomials and inequalities and leverages various algebraic techniques to solve these equations. The study of algebraic approaches to geometry problems has given rise to a range of research achievements, such as Buchberger's Gröbner bases method \cite{buchberger1988applications}, numerical parallel methods \cite{yang1992prover}, polynomial system triangulation elimination algorithm \cite{gao1993dimension}, cylindrical algebraic decomposition for solving inequalities \cite{collins1974quantifier}, dimensionality reduction methods \cite{lu1998practical}, and software tools like GEOTHER \cite{wang2005geother}.

Zhang proposed the point elimination method based on geometric invariants \cite{zhang1995automated}. This approach employs constructive methods to describe problems and is capable of generating concise and meaningful readable proofs for a large number of non-trivial geometric problems. Subsequently, research on machine proofs of geometric theorems based on geometric invariants rapidly advanced \cite{chou1993automated,chou1994collection,li2004symbolic}, leading to the development of practical software tools such as Geometry Explorer \cite{wilson2006geometry}, Geometry Expert \cite{chou1996introduction} and Java Geometry Expert \cite{ye2011introduction}. The method based on geometric invariant can also be extended to solid geometry \cite{chou1995automated} and non-Euclidean geometry \cite{yang1997automated}.

The machine proof of geometric theorems can generally be categorized into the three aforementioned approaches \cite{jingzhong2009automatic}: search-based synthesis methods, algebra methods, and points elimination methods based on geometric invariants. Synthesis method can provide proofs of traditional style but can only prove a small subset of carefully chosen plane geometry theorems due to the limited computational power of computers and the combinatorial explosion inherent in these methods. Algebra methods can handle a wide range of problem types, but the solving process typically involves algebraic expressions that are difficult to manually verify and obtain readable proof procedures. Methods based on geometric invariants can provide readable proof procedures, but the types of problems that can be solved are limited by the types of geometric invariants available.

Geometry problem solving has been gaining more attention in the NLP community recently. Several geometry formal systems and datasets have been constructed, such as Geometry3K \cite{lu2021inter}, GeoQA \cite{chen2021geoqa}, GeometryQA \cite{tsai2021sequence}. Geometry3K translates the known conditions of geometric problems into formal statements, defining theorems as a set of rules for converting between formal statements. This approach, referred to as Formal Language, is also used in GeoRE \cite{yu2021geore}, which focuses on geometric relation extraction, and PGDP5k \cite{hao2022pgdp5k}, which is designed for geometric image parsing. While these methods are intuitive, they lack theoretical guidance, are not comprehensive, and are not easily extensible with additional predicates and theorems. GeoQA employs the formal method of Program, transforming the geometric problem-solving process into a sequence of programs consisting of variables and operators. Executing this program sequence yields the solution. Subsequent work extended the number and types of questions and rules, resulting in GeoQA+ \cite{cao2022augmented}, UniGeo \cite{chen2022unigeo}, and PGPS9K \cite{zhang2023multi}. These formal methods can represent algebraic and symbolic problem-solving processes, but compared to formal language methods, they are less intuitive and cannot represent traditional geometric problem-solving processes. Additionally, adding new rules requires modifying the solver's code, making them less extensible. GeometryQA employs a formal method known as the Expression Tree, which transforms the problem-solving process into a solving tree composed of operators and variables. This method is similar to the programmatic approach but is more structured.

Shared benchmarks and datasets have significantly advanced research in AI-assisted geometric problem solving. Several AI systems, such as CL-based model \cite{jian2023solving}, SCA \cite{ning2023symbolic}, GeoDRL \cite{peng2023geodrl}, have been constructed to achieve higher success rates in problem solving. As problem-solving success rates continue to improve, there is a growing demand for datasets with higher quality and difficulty. Previous work focused on AI system research but overlooked research into geometry formalization theory. Expanding datasets in existing systems requires substantial modifications to solver code, making it challenging to extend both the formal systems and datasets.

\section{Geometry Formalization Theory}

In the realm of geometry, a typical problem consists of known conditions described in natural language, the problem objective, and a geometric diagram. The problem-solving process can be construed as the application of multiple theorems, involving relational reasoning and algebraic computation. The study of GFT focuses on how to transform geometric problems and their solution processes, described in natural language and images, into a unified and precise formal language for mechanical processing by computers. GFT comprises 3 major components: geometry ontology, geometry representation theory, and geometry formal language.

Geometry ontology studies the fundamental ontology within the field of geometry and the relationships between these ontological elements. It employs the geometry ontology domain, which provides a comprehensive and systematic summary of the knowledge of Euclidean plane geometry. The geometry ontology domain further refines into the geometry knowledge graph, guiding the design of geometry formal systems.

Geometry representation theory investigates how to express geometric knowledge using formal language. Formal systems are abstract descriptions and simulations of the real world, with a one-to-one correspondence to the real world. Consistency theory explores how to establish a correct formal system. Under the guidance of consistency theory, we propose geometry representation theory. When transforming various geometric knowledge into a unified formal language, we must ensure the consistency of static representations (such as the formal representation of diagrams) and the consistency of dynamic processes (such as theorems described in formal language).

Geometry formal language takes the form of structured geometric knowledge, divided into two categories: geometry definition language (GDL) and condition declaration language (CDL). GDL includes predicate definition language and theorem definition language. The former defines various types of geometric relations and properties, while the latter defines various theorems that may be used in the problem-solving process. During the initialization phase of a geometric problem solver, GDL is used to configure the solver, achieving its shareability and extensibility. CDL is employed to describe the known conditions of individual geometric problems and is divided into three categories: construction statements, condition statements, and objective statements. CDL allows us to input geometric problems into the solver.

\subsection{Geometry Ontology}

Geometry ontology domain consists of two dimensions: number and shape, and static and dynamic. Number refers to precise and quantitative descriptions of geometric knowledge, while shape pertains to generalized and qualitative descriptions of geometric knowledge. Static refers to various geometric knowledge elements, such as the properties of geometric figures and their interrelationships. Dynamic encompasses rules for transforming different types of geometric knowledge, including common knowledge and theorems.

\begin{figure}
\centering
\includegraphics[width=0.8\textwidth]{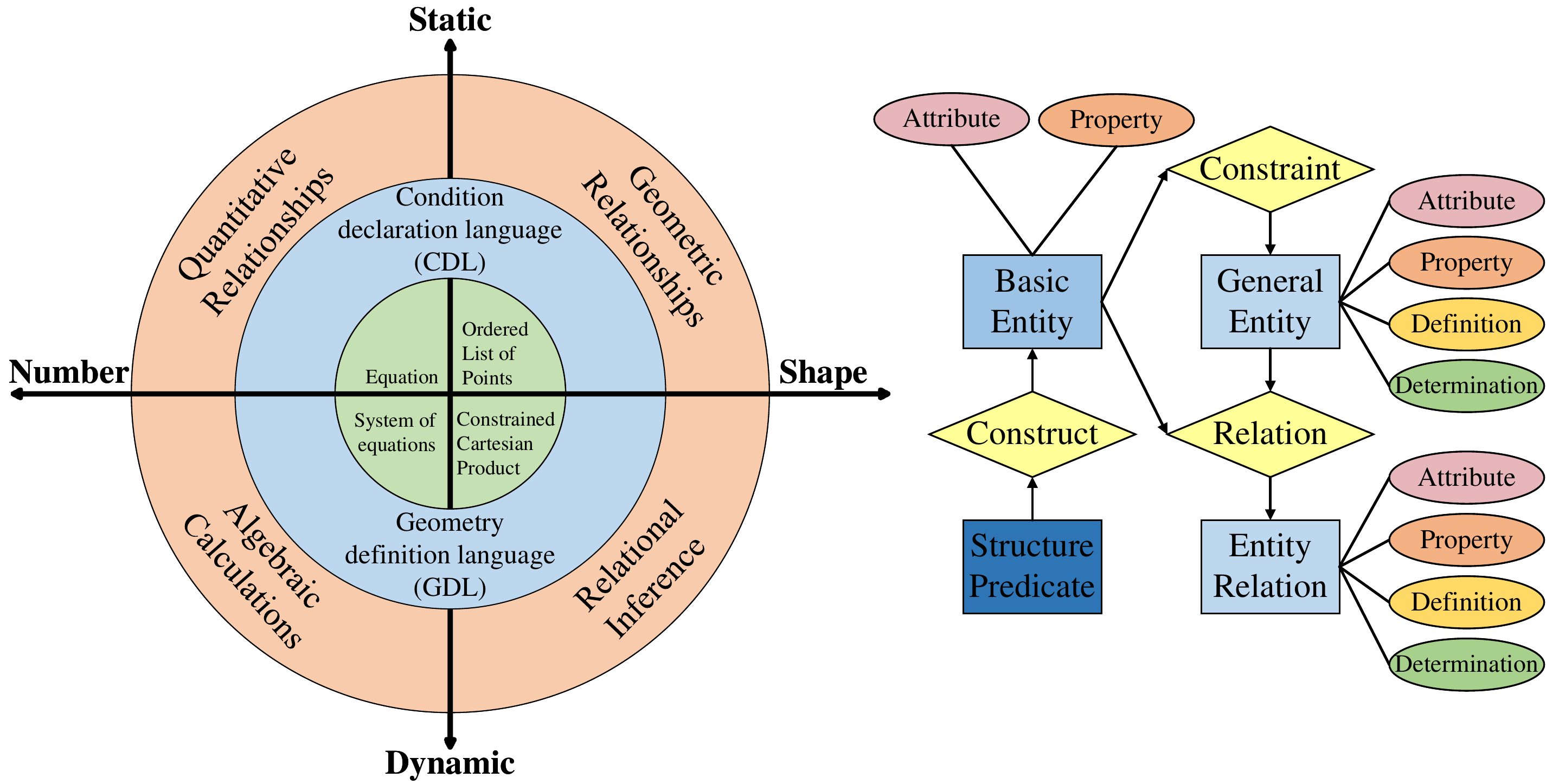}
\caption{Geometry ontology domain (left). It consists of two dimensions: number and shape, and static and dynamic and further subdivided into three levels: axiomatic systems, formal systems, and solvers. Simplified geometry knowledge graph (right). Rectangles represent static geometric knowledge, while circles represent dynamic processes of transforming geometric knowledge. An example can be found in App.~\ref{app-GKG}.}
\label{fig-geometry_ontology_domain}
\end{figure}

These two dimensions divide the geometry ontology domain into four quadrants, each further subdivided into three levels of mapping relationships: axiomatic systems, formal systems, and solvers, as shown in Fig.~\ref{fig-geometry_ontology_domain}. The outermost layer, axiomatic systems, refers to geometric knowledge described in natural language or diagrams, such as problem conditions and theorem definitions. The intermediate layer, Formal Systems, transforms vague and uncertain natural language descriptions and knowledge in different forms from text and diagrams into precise, human-readable, and computable formal languages, serving as a bridge between humans and computers. The innermost layer, Solvers, represents the specific internal form of geometric knowledge within a computer, including data structures for problem conditions and applying methods for theorems. App.~\ref{app-GOD} provides examples for various components of the geometry ontology domain to facilitate understanding.

By further expanding geometry ontology domain, we construct geometry knowledge graph as shown in Fig.~\ref{fig-geometry_ontology_domain}. Construction statements contain all the structural information of geometric figures. Starting from three types of construction statements, we derive all basic entities. By imposing further constraints on basic entities, we obtain general entities. Basic entities and general entities interact internally and with each other, forming entity relationships. Attributes represent quantifiable descriptions of geometric objects. Construction statements, basic entities, general entities, entity relationships, and attributes describe the static aspects of geometric knowledge. Properties, definitions, and judgments describe the dynamic processes of transforming geometric knowledge, i.e., theorems. The knowledge graph of geometry provides a detailed representation of the relationships and hierarchical structure among various geometric knowledge components. Using the knowledge graph of geometry ensures that the constructed formal system is more comprehensive and avoids omissions.

\subsection{Geometry Representation Theory}

App.~\ref{app-consistency_theory} introduces the consistency theory of formal systems. Guided by this theory, when constructing a geometry formal system, we must ensure the consistency of static representation and the consistency of dynamic processes.

\subsubsection{Consistency in Static Representation}

In the field of geometry, various geometric knowledge is conveyed through textual descriptions or geometric diagrams. Geometric knowledge in textual form is presented using natural language and mathematical symbols, making it relatively straightforward to transform into structured formal language. However, the challenge lies in formalizing the geometric knowledge implicit in geometric diagrams, which necessitates establishing a reversible mapping between geometric diagrams and their formal representations.

We classify the information inherent in geometric diagrams into two categories: topological structure information (TSI) and metric information (MI). TSI defines the fundamental structure of a diagram and serves as a crucial basis for classifying different geometric shapes. MI further characterizes various properties of the diagram, such as the length of lines and the measure of angles. MI is closely related to TSI and is relatively amenable to summarization and formalization. The primary challenge lies in formalizing the TSI.

The most fundamental elements that compose a geometric diagram are points, which can be used to describe the TSI of the diagram. We can define a set of rules that utilize the points constituting the diagram to depict its TSI.

\begin{figure}
\centering
\includegraphics[width=\textwidth]{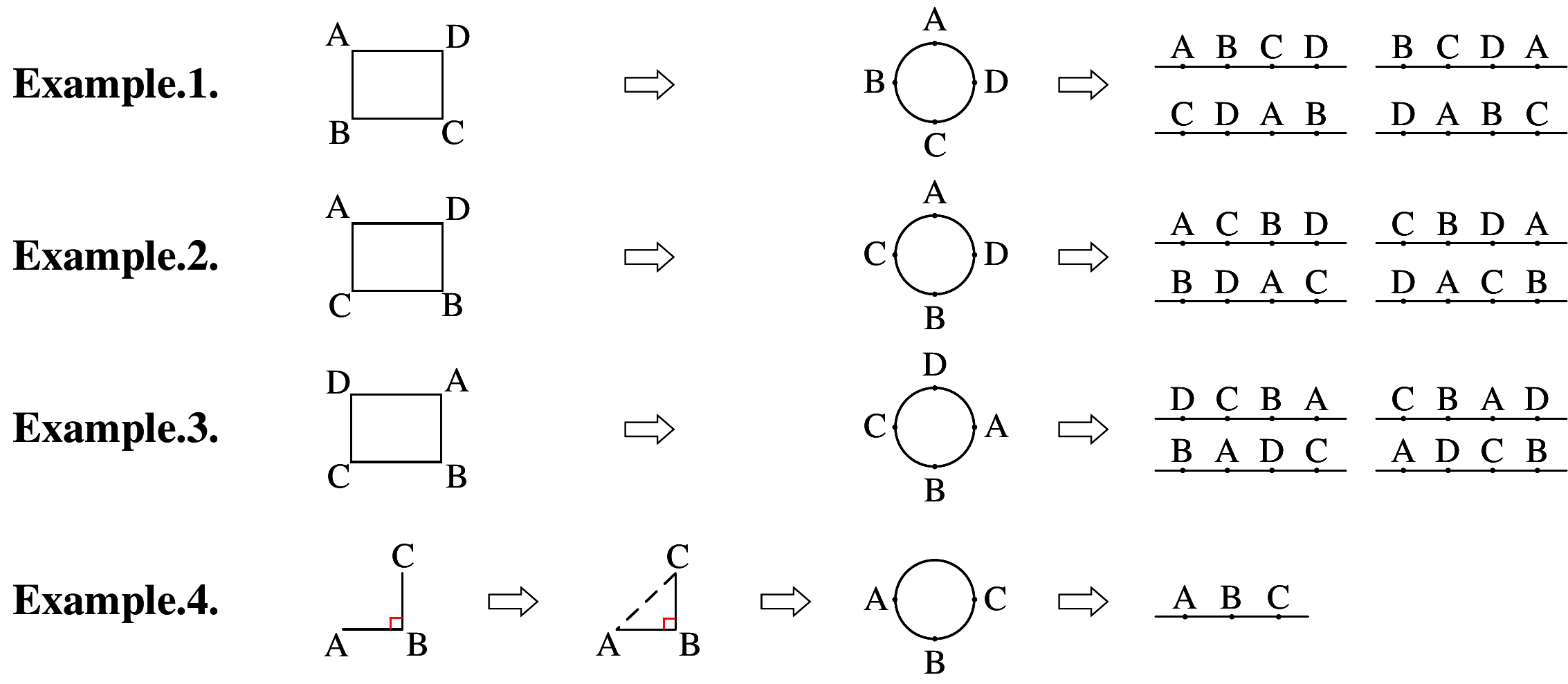}
\caption{Examples of topological mapping method}
\label{fig-topological_mapping_method}
\end{figure}

Closed geometric figures can be transformed into topologically equivalent circles, as shown in Fig.~\ref{fig-topological_mapping_method}. Unfolding the circle from any position of the topologically equivalent circle into a straight line, the relative positions of points on the line record the topological structural information of the original diagram. We can use an ordered list of points as the formal representation of the geometric topological structural information. For non-closed geometric figures, we can first transform them into closed geometric diagrams before proceeding with formalization, as shown in Example.4. of Fig.~\ref{fig-topological_mapping_method}. By unfolding the topologically equivalent circle from different positions, we may obtain different ordered lists of points. All these ordered lists together form a set, serving as the formal representation of the topological structural information of geometric diagram. Any element in the set contains all the topological structural information of the original diagram.

The basic transformations of a geometric diagram can be defined as several operations on its formal representation, as shown in App.~\ref{app-TMM}.

The above method is called the topological mapping method, which can be used for the formalization of a simple geometric diagram. In practice, a geometric problem's diagram is often composed of a combination of simple geometric diagrams. We propose the topological construction method, which utilizes the formal representations of multiple simple geometric diagrams to obtain the formal representation of a composite diagram.

\begin{figure}
\centering
\includegraphics[width=\textwidth]{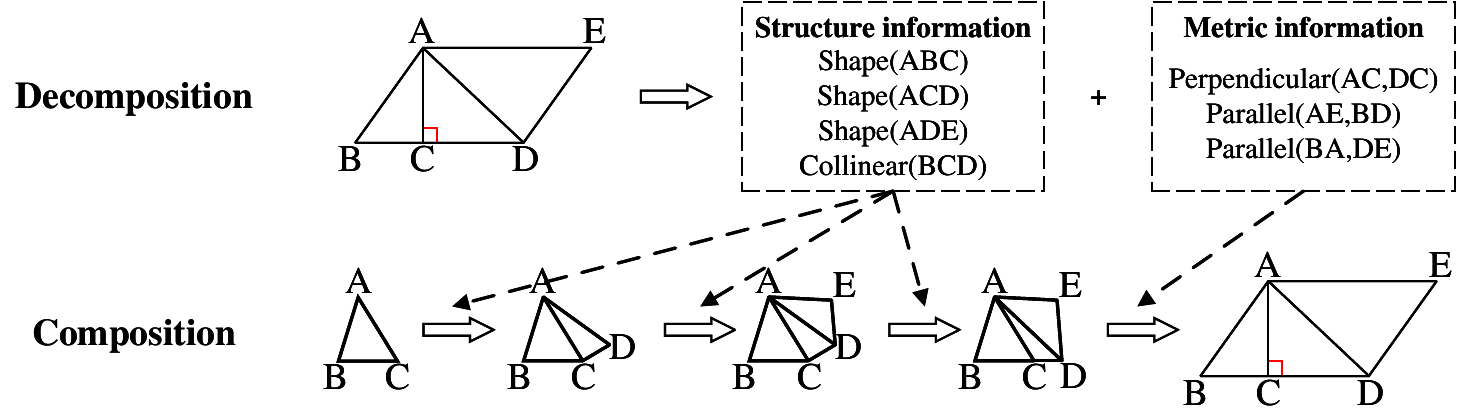}
\caption{An example of topological construction method}
\label{fig-topological_construction_method}
\end{figure}

The topological construction method decomposes the composite diagram into simple geometric diagrams while preserving the TSI between them, TSI of each simple geometric diagram and MI of each simple geometric diagram. When constructing a diagram, it first reconstructs the basic structure of the original diagram based on the TSI and then adjusts the diagram further according to the MI to obtain the original diagram. The topological construction method is a constructive drawing method that is not affected by the order of construction statements, as shown in Fig.~\ref{fig-topological_construction_method}. We have implemented topological construction method in FGPS, and the algorithm description and time complexity analysis can be found in App.~\ref{app-TCM}.

The process of constructing a geometric diagram can be denoted as $\oplus$. If diagram $C$ is composed of diagrams $A$ and $B$, their formal representations of TSI are represented as sets $R_a$, $R_b$, and $R_c$, respectively. Diagram $A$ contains $m$ points, while diagram $B$ contains $n$ points. There are $k$ ($k\geq2$)common points shared between $A$ and $B$. Two elements $P_a=(p^{(a)}_1, \dots, p^{(a)}_{s_1}, p_i, \dots, p_{i+k}, p^{(a)}_{s_2}, \dots, p^{(a)}_m)$ and $P_a=(p^{(b)}_1, \dots, p^{(b)}_{s_1}, p_{i+k}, \dots, p_i, p^{(b)}_{s_2}, \dots, p^{(b)}_n)$ are selected from sets $R_A$ and $R_B$. $\oplus$ is defined in two steps, as shown in Eq.~\ref{eq-TCM1} and Eq.~\ref{eq-TCM2}. In the first step, $\otimes$ operation is applied to $P_a$ and $P_b$ to obtain the element $P_c$. In the second step, the $P_c$ undergoes the $rotate$ (Eq.\ref{eq-rotate}) $i$ times to construct the set $R_c$.

\begin{equation}
\begin{aligned}
    &P_a \otimes P_b \\
    =&(p^{(a)}_1, \dots, p^{(a)}_{s_1}, p_i, p^{(b)}_{s_2}, \dots, p^{(b)}_n, p^{(b)}_1, \dots, p^{(b)}_{s_1}, p_{i+k})
\end{aligned}
\label{eq-TCM1}
\end{equation}

\begin{equation}
\begin{aligned}
    &R_a \oplus R_b\\
    =&\{rotate^{(i)}(P_a \otimes P_b)|i=1,2, \dots, m+n-2k+2\}\\
    =&R_c
\end{aligned}
\label{eq-TCM2}
\end{equation}

The formal representation of composite diagrams, along with $\oplus$, forms a semigroup, satisfying closure, commutative law and associative law. The proof can be found in App.~\ref{app-TCM}.

\subsubsection{Consistency in Dynamic Process}

In essence, geometric theorems are rules governing the transformation of various geometric knowledge, involving relation reasoning, logical operations, and algebraic calculations. Geometric predicate logic (GPL) translates the diverse operations inherent in geometric theorems into a unified formal language, enabling precise theorem descriptions that can be mechanically executed by computers.

Geometric knowledge comprises geometric relations and quantitative relations, with geometric relations as the core, and quantitative relations are attributes of geometric relations. Multiple geometric relations can lead to new relations through relation reasoning, where elements in these new relations can be derived through operations on the elements of existing relations.

\begin{figure}
\centering
\includegraphics[width=\textwidth]{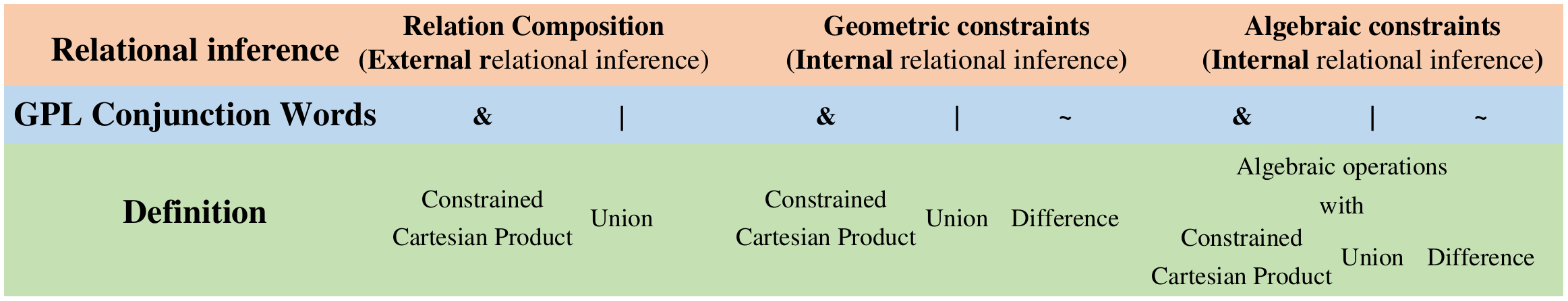}
\caption{Definition of geometric predicate logic}
\label{fig-geometric_predicate_logic}
\end{figure}

GPL categorizes operations in geometry into external relation reasoning and internal relation reasoning, as illustrated in Fig.~\ref{fig-geometric_predicate_logic}. If the operation results in a new relation with a different structure, it is termed external relation reasoning, also known as relation composition. Confining a relation with constraints to obtain a stricter relation while preserving the original relation's structure constitutes internal relation reasoning. Depending on the type of constraint applied, this can be further categorized into geometric constraints and algebraic constraints on relations. Relation reasoning is subdivided into basic operations $\&$, $|$, and $\sim$, which correspond to logical operations AND, OR, and NOT, respectively.

Geometric relations are denoted as $R(v_1, v_2, \dots, v_n), v \in V$, where $R$ is referred to as the relation name, specifying the type of geometric relation, and $V$ represents point variables, describing the TSI of the geometric relation. A geometric relation with $N$ elements is represented as a set $R(v_1, v_2, \dots, v_n) = \{ (p^{(i)}_1,p^{(i)}_2, \dots, p^{(i)}_n) | i = 1, 2, \dots, N\}$, where $p$ are the points constituting the geometric relation. For any element $r_i = (p^{(i)}_1,p^{(i)}_2, \dots, p^{(i)}_n)$ of a geometric relation, we define $r_i(v_j)$ as the value of the position $v_j$ of element $r_i$, i.e., $r_i(v_j) = p_j$. Quantity relations can be represented by algebraic constraints denoted as $R_A$, where $R_A(r) = 1$ indicates that an element satisfies the constraint. Given relations $R_1(v^{(1)}_1, v^{(1)}_2, \dots, v^{(1)}_n)$ and $R_2(v^{(2)}_1, v^{(2)}_2, \dots, v^{(2)}_m)$, they have $k$ common variables and their point variables are denoted as $V_1$ and $V_2$. We can obtain a new relation $R_3$ through GPL.

$\&$ is referred to as the constrained Cartesian product and is analogous to the logical operation AND. The operation is denoted as $R_1 \& R_2 \rightarrow R_3$. Initially, the Cartesian product operation is applied to $R_1$ and $R_2$, yielding $R^{'}_3$. Constraints are imposed on $R^{'}_3$ to select elements that adhere to the constraint conditions. These constraints necessitate that the elements within $R^{'}_3$ exhibit identical values at the common variable positions of $R_1$ and $R_2$, as shown in Eq.~\ref{eq-GPL1}, where $\times$ represents the Cartesian product. Then, duplicate common variables are eliminated, leading to the derivation of a new geometric relation $R_3$, as shown in Eq.~\ref{eq-GPL2}.

\begin{equation}
\begin{aligned}
    &R^{'}_3(v^{(1)}_1, v^{(1)}_2, \dots, v^{(1)}_n, v^{(2)}_1, v^{(2)}_2, \dots, v^{(2)}_m)\\
    =&\{(r^{1}_i, r^{2}_j) | (r^{1}_i, r^{2}_j) \in R_1 \times R_2, r^{1}_i(v) = r^{2}_j(v), v \in V_1 \cap V_2 \}
\end{aligned}
\label{eq-GPL1}
\end{equation}

\begin{equation}
\begin{aligned}
    &R_1(v^{(1)}_1, v^{(1)}_2, \dots, v^{(1)}_n) \& R_2(v^{(2)}_1, v^{(2)}_2, \dots, v^{(2)}_m)\\
    =&\{(r(v^{(3)}_1), r(v^{(3)}_2), \dots, r(v^{(3)}_{m+n-k})) | r \in R^{'}_3, v^{(3)} \in V_1 \cup V_2\} \\
    =&R_3(v^{(3)}_1, v^{(3)}_2, \dots, v^{(3)}_{m+n-k})
\end{aligned}
\label{eq-GPL2}
\end{equation}

The definition of $\&$ is more straightforward when the second relation is a quantitative relation. The operation is denoted as $R_1 \& R_A \rightarrow R_3$. The point variables of $R_A$ are a subset of those in $R_1$. We take the elements from $R_1$ and incorporate them into $R_A$ corresponding to the point variables to construct algebraic constraints. The set of all elements that satisfy these algebraic constraints forms a new geometric relation $R_3$, as shown in Eq.~\ref{eq-GPL3}.

\begin{equation}
R_3(v^{(1)}_1, v^{(1)}_2, \dots, v^{(1)}_n) =\{r | r \in R_1, R_A(r) = 1\} 
\label{eq-GPL3}
\end{equation}

$|$ corresponds to the logical operation OR, represented as $R_1 | R_2 \rightarrow R_3$, as shown in Eq.~\ref{eq-GPL4}. $|$ is commonly nested together with $\&$ in practical applications. 

\begin{equation}
R_3(v^{(1)}_1, v^{(1)}_2, \dots, v^{(1)}_n) =\{r | r \in R_1 \cup R_2\} 
\label{eq-GPL4}
\end{equation}

$\sim$ corresponds to the logical operation NOT, represented as $\sim R_1 \rightarrow R_3$, as shown in Eq.~\ref{eq-GPL5}. $E$ is defined as all possible elements within a certain relation, specifically, all permutations of known points that conform to the structure of $V_1$.

\begin{equation}
R_3(v^{(1)}_1, v^{(1)}_2, \dots, v^{(1)}_n) =\{r | r \in E - R_1\} 
\label{eq-GPL5}
\end{equation}

GPL satisfies commutative, associative, and distributive laws, as demonstrated in App.~\ref{app-GPL}. The nested use of GPL connectives with geometric and numeric relations provides a powerful expressive capability and can be employed for the formalization of theorems, as illustrated in App.~\ref{app-TDL}.

\subsection{Geometry Formal Language}

Formal languages are categorized into geometry definition language (GDL) and conditional declaration language (CDL). The former is used to define entities, attributes, theorems, and other elements of a geometry formal system, while the latter is employed for declaring known conditions and problem-solving objectives in geometric problems.

GDL comprises predicate definition language and theorem definition language. Predicate definition language is used to define different types of geometric relations and geometric properties. A typical predicate definition statement includes the name of the geometric relation, point variables, existential constraints on entities, format validity constraints, and automatic extension. The solver can read and interpret predicate definition language to ensure the legitimacy of input problem conditions.

CDL consists of three main parts: construction statements, condition statements, and goal statements. Construction statements describe the TSI of the geometric problem's diagram, such as basic shapes, collinear, and cocircular. Condition statements are used to input the known conditions in geometric problems, including both geometric and algebraic relations. goal statements declare the problem-solving objectives.

Both types of formal languages share the same syntax format, which is similar to predicate logic syntax. The fundamental concepts are predicates and terms. Predicates are used to define categories of geometric knowledge, encompassing geometric relations and algebraic relations. Terms specify the specific content of this knowledge. If it's a geometric relation, the term is an ordered sequence of points; if it's an algebraic relation, the term is an expression composed of geometric attributes, operators, free variables, and numbers. Functions are mappings from individual geometric relation terms to individual algebraic relation terms. Through such mappings, ordered sequences of points can be used to represent numeric relations, unifying the representation formats of geometric and numeric relations. Examples of formal language can be found in App.~\ref{app-PDL}, App.~\ref{app-TDL}, and App.~\ref{app-CDL}.

\section{Geometry Formal System: FormalGeo}

Guided by GFT, we have constructed the geometry formal system, FormalGeo. This formal system comprises 88 predicates and 196 theorems (see App.~\ref{app-PDL} and App.~\ref{app-TDL}) and can represent, verify, and solve geometric problems ranging from SAT-level to IMO-level. Based on FormalGeo, we annotated the formalgeo7k and formalgeo-imo datasets.

The geometric definition language serves as the specific form of the geometry formal system. The design of the geometry formal system includes the design of the predicate definition language and the theorem definition language. This section introduces the design methodology of FormalGeo and provides an overview of formalgeo7k and formalgeo-imo datasets.

Utilizing the simplified geometric knowledge graph presented in Fig.~\ref{fig-geometry_ontology_domain} as a template, we formulate the predicate definition language and theorem definition language for FormalGeo.

\subsection{Predicate Definition Language}

Structure predicate are used to describe the TSI of geometric figures. FormalGeo comprises three fundamental structure predicates: Shape, Collinear, and Cocircular. In the problem-solving initialization phase, the solver executes topological construction method based on recognized structure predicate statements, automatically expanding to 6 basic entities: Point, Line, Angle, Polygon, Arc and Circle. Basic entities are a more intuitive representation of the TSI of geometric figures and serve as the core for various geometric and algebraic relations. Structure predicate and basic entity together are referred to as construction predicate, which describe the TSI of the figures.

\begin{figure}
\centering
\includegraphics[width=0.5\textwidth]{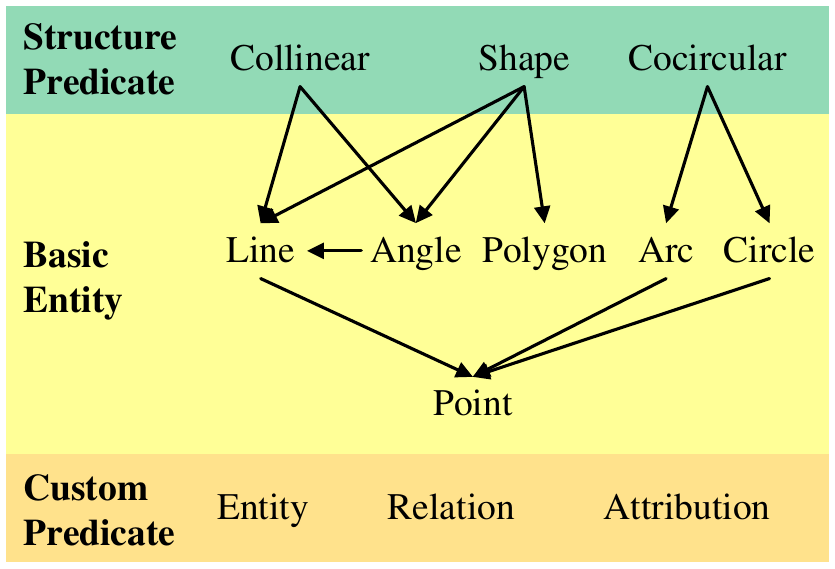}
\caption{Hierarchy of predicates}
\label{fig-predicates_hierarchy}
\end{figure}

By applying additional constraints to basic entities, general entities are obtained. Various geometric entities are interconnected, forming entity relations. Properties are used to describe the MI of entities and entity relations, in conjunction with free variables, operators, and real numbers, to establish quantitative relations. General entity, entity relation, and property are collectively referred to as custom predicate. Fig.~\ref{fig-predicates_hierarchy} illustrates the hierarchical structure and extension relationships among various geometric predicates. The solver can automatically extend conditions based on the extension rules defined in the predicate definition language. It should be noted that the principle of extension rules is followed that construction predicate are only extended by other construction predicate, and custom predicate are only extended by other custom predicate.

\subsection{Theorem Definition Language}

Geometric theorems in a formal system are described using geometric predicate logic, consisting of premises and conclusions. Based on the characteristics of their premises and conclusions, geometric theorems can be broadly categorized into three types: properties, definitions, and determinations. If the premises of a theorem contain only one geometric relation, the theorem falls into the category of properties or definitions. Definitions are considered common knowledge and are automatically invoked and extended by the solver, while properties require explicit invocation. If the conclusion of a theorem contains only one geometric relation, the theorem serves as a determination for that particular geometric relation.

\begin{figure}
\centering
\includegraphics[width=0.9\textwidth]{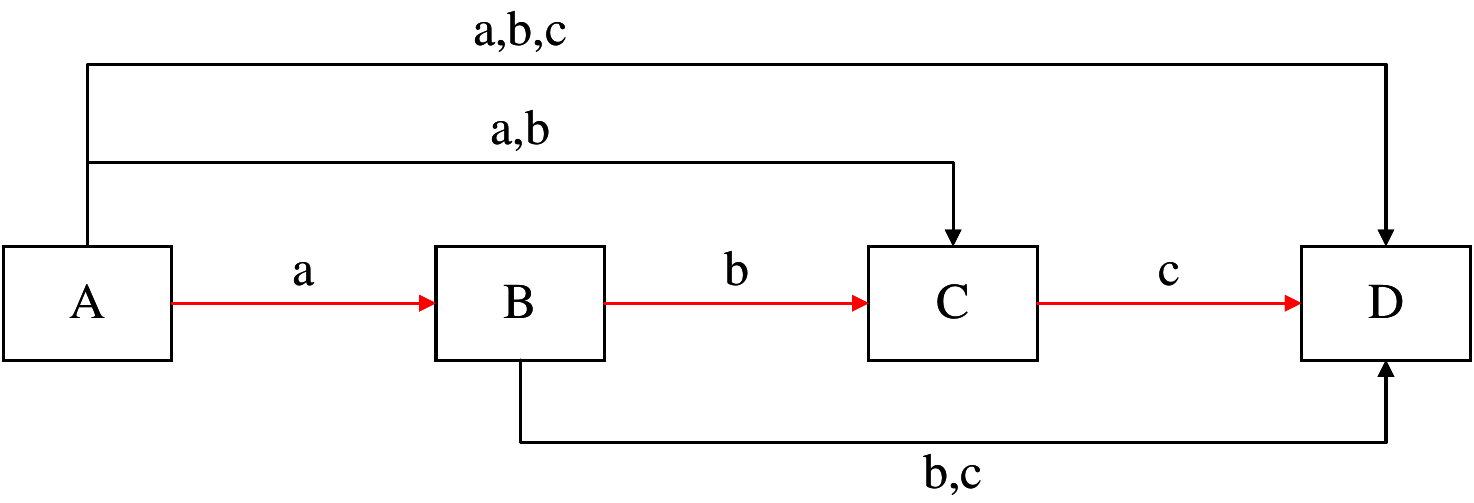}
\caption{Hierarchy of theorems}
\label{fig-theorems_hierarchy}
\end{figure}

As previously mentioned, geometric theorems are rules for converting various geometric knowledge, and these rules exhibit a hierarchical structure, as depicted in Fig.~\ref{fig-theorems_hierarchy}, where each arrow represents a method of theorem definition. When defining theorems, we can directly combine multiple predicates to create a determination theorem, as shown in Eq.~\ref{eq-TDL1}, or we can break it down into several theorems, as shown in Eq.~\ref{eq-TDL2}. Different ways of defining theorems can impact the speed of theorem application in the problem-solving process. Therefore, it is essential to explore the most suitable method for defining theorems.

\begin{equation}
    A \& a \& b \& c \rightarrow D
\label{eq-TDL1}
\end{equation}

\begin{equation}
A \& a \rightarrow B, B \& b \rightarrow C, C \& c \rightarrow D
\label{eq-TDL2}
\end{equation}

We introduce the concept of the abstract hierarchy of theorems to describe the level of structure in theorem definitions. The abstract hierarchy of a theorem, denoted as $K$, represents the minimum number of theorems required to derive a higher-level predicate from lower-level predicates. Along the red paths in Fig.~\ref{fig-theorems_hierarchy}, we have $K_{A \rightarrow B} = 1$, $K_{A \rightarrow C} = 2$, and $K_{A \rightarrow D} = 3$.

Theorems are intended to facilitate the solving of problems. We prioritize theorem definition based on solving time, while temporarily disregarding other factors such as readability. For search-based problem-solving algorithms, the solving time $T$ is influenced by the length of the theorem sequence $d$, the number of theorems $N$ in theorem library, and the average application time $\bar{t}$ of a single theorem. This can be defined more specifically in Eq.~\ref{eq-TDL3}. Here, $K$ is directly proportional to $d$ and inversely proportional to $N$ and $\bar{t}$.

\begin{equation}
T \propto f(d, N, \bar{t}) \propto (N + N^2 + \dots + N^d)\bar{t}
\label{eq-TDL3}
\end{equation}

In our practical observations, we have found that there exists an approximate inverse relationship between $T$ and $K$. In the process of defining theorems, FormalGeo tends to favor higher abstract hierarchy. We leave more detailed comparative experiments and mechanistic analyses for future work.

\subsection{Datasets}

Most of the existing datasets for geometry problem-solving suffer from the following issues: 1.Limited data volume or non-open source availability. 2.Lack of annotations or incomplete and low-quality annotations. 3.Absence of formalazation theory support, resulting in incoherent and inconsistent formal systems. 4.Low scalability,  Defining new predicates and theorems require solver's code modifications. 5.Lower difficulty level of the problems. To address the aforementioned issues, we have annotated formalgeo7k and formalgeo-imo.

Our data is collected from various sources, including Geometry3k \cite{lu2021inter}, GeoQA \cite{chen2021geoqa}, GeoQA+ \cite{cao2022augmented}, and online resources. We carefully curated, classified, deduplicated, and standardized the problem statements. The creation of the formalgeo7k involved 16 trained master's students over a period of around 13 weeks. The creation of the formalgeo-imo involved 4 trained master's students over a period of around 1 week. Excluding the time spent on collaboration and dataset allocation, annotating datasets took approximately 1000 person-hours. 

formalgeo7k comprises 6981 geometric problems that are accompanied by natural language descriptions, geometric diagrams, formal language annotations, and solution theorem sequence annotations. A annotated problem is illustrated in in Fig.~\ref{fig-problem_example}. The problem-solving process can be represented as a hypertree with conditions as hypernodes and theorems as hyperedges. The solution theorem sequence is a path from the root node (known conditions) to a leaf node (the problem-solving objective). By selecting any intermediate node along this path as the problem-solving objective, we can generate new problems, allowing us to expand the problem number to 133,818. formalgeo-imo is constructed with the same standards but with more challenging problem difficulty.

\begin{figure}
\centering
\includegraphics[width=\textwidth]{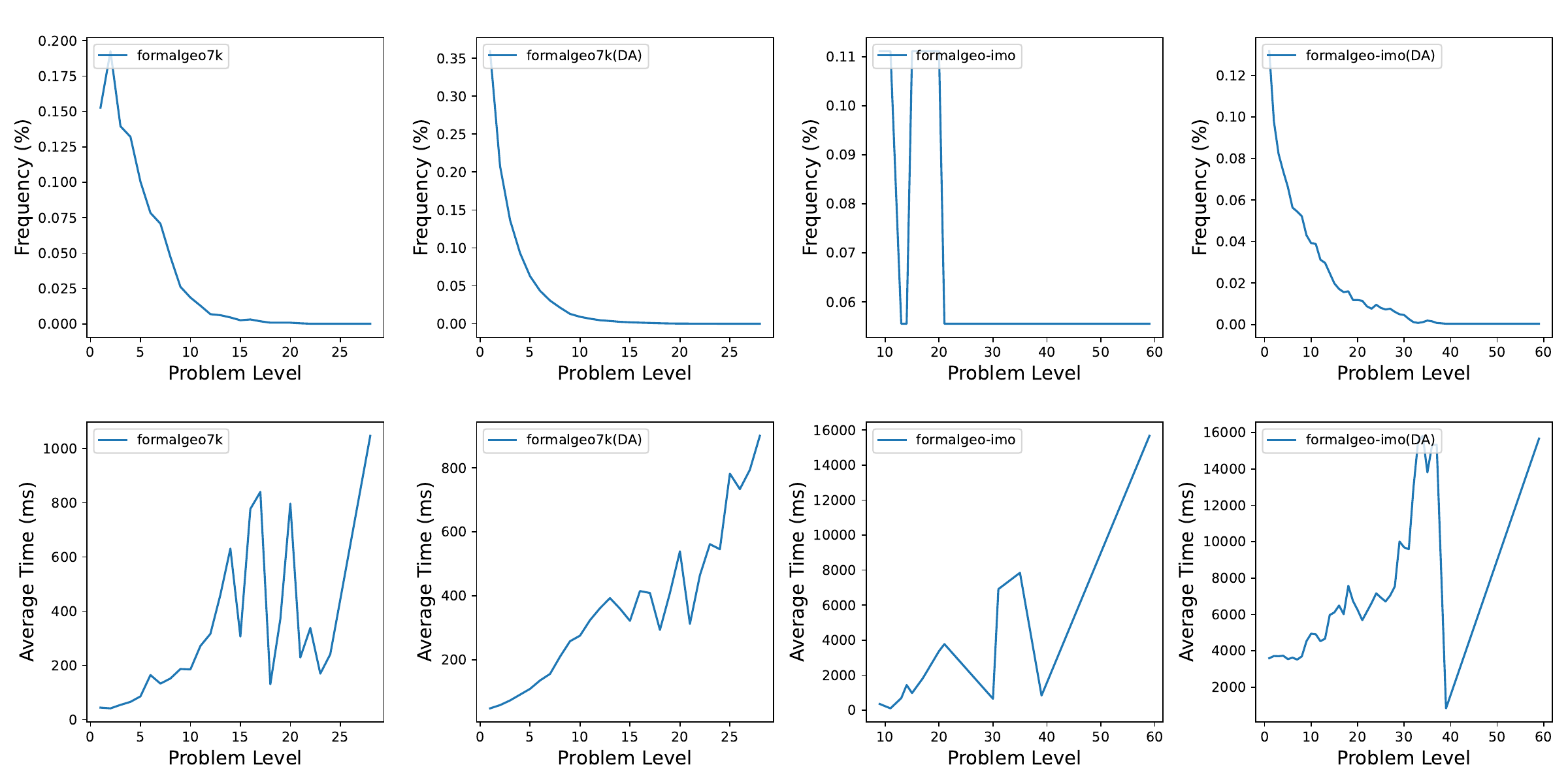}
\caption{Distribution of problem (the top 4). Average Time of interactive verification (the bottom 4). DA represents data augmentation.}
\label{fig-chart_problem_level}
\end{figure}

We use the length of theorem sequences required for problem solving as a rough metric for assessing problem difficulty. All annotated and expanded problems have been verified by the solver, and their average solution times varying with problem difficulty also show in the Fig.~\ref{fig-chart_problem_level}. The number of questions with a difficulty level of 15 or higher in formalgeo7k is quite small, leading to significant fluctuations. formalgeo-imo follows the same reasoning. After data augmentation, datasets exhibit a larger scale of data and a smoother difficulty curve. In general, more challenging problems require longer solving time. 

\section{Geometry Problem Solver: FGPS}

Guided by GFT, we have constructed a geometric problem solver, FGPS. Any geometric formal system designed based on GFT and any geometric problem that adheres to the syntax of formal language can be input into FGPS for verification and solution.

In this section, we introduce the implementation of the core solving engine. Detailed descriptions about solver's structure and other functionalities can be found in the App.~\ref{app-FGPS}.

\subsection{GPL Executor}

The process of geometric problem solving can be represented as a sequence of theorem applications. Theorems are defined using GPL, and, as a result, the process of geometric problem solving within the solver is essentially the execution of GPL. GPL statements can consist of multiple logical conjunction words, geometric relations, and quantitative relations nested together. The application process can be divided into 4 steps:

In the GPL parsing phase, the solver expands complex GPL statements into disjunctive normal form (DNF) using the distributive law. Each simple conjunction represents a branch of the theorem. This not only meets the requirements for backward reasoning and facilitates the generation of sub-goals but also speeds up theorem execution by skipping irrelevant branches.

In the GPL ordering phase, for each branch of the theorem, the solver adjusts the positions of geometric relations and quantitative relations within simple conjunctions according to the commutative law. The guiding principles for this adjustment are as follows: 1.Transforming relation composition into geometric constraints 2.Moving geometric constraints forward 3.Moving algebraic constraints backward. This approach not only helps filter out geometric relation elements that do not comply with the constraints, preventing the explosion of combinatorics caused by Cartesian product operations, but also reduces the number for algebraic equation solving, thereby improving theorem application speed.

In the GPL execution phase, the solver reads geometric and quantitative relations sequentially and performs relational inference (Eq.~\ref{eq-GPL1} $\sim$ ~\ref{eq-GPL5}) in the order of their appearance.

The GPL execution process can be illustrated with an example. Suppose that we have a theorem defined as shown in Eq.~\ref{eq-gple1}, which includes 5 geometric relations $R_1(v_1, v_2)$, $R_2(v_2, v_3)$, $R_3(v_2)$, $R_4(v_2, v_3)$ and $R_5(v_2)$ and 1 quantitative relation $R_A(v_1, v_2)$.

\begin{equation}
R_1 \& (R_2 | ( \sim R_3 | R_A) \& R_4 \& R_5)  
\label{eq-gple1}
\end{equation}

During the GDL parsing phase, it is expanded into a DNF according to the distributive law, as shown in Eq.~\ref{eq-gple2}. This DNF consists of 3 simple conjunctions, with each simple conjunction serving as a theorem branch.

\begin{equation}
R_1 \& R_2 | R_1 \& \sim R_3 \& R_4 \& R_5 | R_1 \& R_A \& R_4 \& R_5
\label{eq-gple2}
\end{equation}

In the GDL reordering phase, let's take branch $R_1 \& R_A \& R_4 \& R_5$ as an example. It adjusts the order of its statements according to the commutative law, resulting in the form shown in Eq.~\ref{eq-gple3}.

\begin{equation}
R_1 \& R_5 \& R_4 \& R_A
\label{eq-gple3}
\end{equation}

In the GPL execution phase, the GDL statements are read and executed in order, and the process is as shown in Eq.~\ref{eq-gple4}.

\begin{equation}
R_1 \& R_5 \& R_4 \& R_A \rightarrow R_{1,5} \& R_4 \& R_A \rightarrow R_{1,5,4} \& R_A \rightarrow R_{1,5,4,A}
\label{eq-gple4}
\end{equation}

\subsection{Minimum Dependency Equations}
The known conditions of geometric problems can be categorized into geometric and quantitative relationships. Quantitative relationships eventually represented as a set of algebraic equations or inequalities. When performing algebraic constraint in the execution of GPL, the satisfaction of algebraic constraint under the known algebraic equations or inequalities of the problem is checked.

Algebraic constraints can be transformed into algebraic expressions represented by $a$, creating the target equation $g-a$. Among the several known equations $X$ in the problem conditions, those relevant to $g-a$ are selected to construct the target equation group $G$, which is subsequently solved. If $g=0$ is obtained as a solution, the algebraic constraints are satisfied. Typically, only a few equations in $X$ are related to $g-a$, and this subset of equations is referred to as the minimum dependency equations.

The solving of equations accounts for the majority of the time spent in the entire process of solving geometric problems. Accelerating the equation solving process is crucial for enhancing the speed of geometric problem solving. To address this, we propose a method for constructing the minimum dependency equations. Without loss of generality, we examine the intermediate process of constructing $G$. At time $t$ ($t=1,2,\ldots$), $G_t$ contains $t$ equations and $m$ unknowns, with the set of unknowns denoted as $M_t$. We need to select a candidate equation $x_t$ from $X$ to add to $G_t$ in a way that increases the likelihood of obtaining a solution for the unknown $g$. The set of unknowns in $x_t$ is represented as $B_t$. This process is repeated until $|M_t|=t$ or no new equations can be added.

\begin{equation}
|B_t \cap M_t| > 0
\label{eq-minieq1}
\end{equation}

\begin{equation}
min(|B_t - M_t|)
\label{eq-minieq2}
\end{equation}

\begin{equation}
max(|B_t \cap M_t|)
\label{eq-minieq3}
\end{equation}

The selection criteria for $x_t$ are as follows:

1.$B_t$ must intersect with $M_t$, as shown in Eq.~\ref{eq-minieq1}. If they do not intersect, it implies that $x_t$ is unrelated to $G_t$.

2.Under the condition of satisfying Eq.~\ref{eq-minieq1}, adding $x_t$ should introduce as few new unknown variables as possible, as depicted in Eq.~\ref{eq-minieq2}. The closer the number of $t$ and $|M_t|$ are, the higher the likelihood of solving $G_t$. In the initial stages of constructing $G$, which only contains $g-a$, $M_1 - 1 > 1$. The number of Added equation each time is a fixed value 1. If we aim to minimize the gap between $t$ and $|M_t|$, we should try to introduce as few new equations when selecting $x_t$.

3.Under the condition of satisfying Eq.~\ref{eq-minieq1} and Eq.~\ref{eq-minieq2}, the equation to be added should encompass more unknown variables, as demonstrated in Eq.~\ref{eq-minieq3}. These additional unknown variables are often associated with other equations within $G_t$, providing more choices for simplifying $G_t$. If there are multiple equations that satisfy these conditions, we can choose any of them at random.

\section{Experiments}

We conducted experiments on the formalgeo7k, comparing different search methods and strategies in terms of problem-solving success rate, solution time, and the number of steps required for problem-solving.

Forward search (FW) starts from the known conditions of the problem and continuously apply theorems to derive new conditions until the goal is achieved. Backward search (BW), on the other hand, begins with the problem-solving goal, expands it into multiple sub-goals, and repeats this process until all sub-goals are resolved. A detailed description of the search algorithms can be found in App.~\ref{app-FGPS}.

The search-based methods construct a search tree during the problem-solving process. We have the flexibility to choose various strategies to traverse the search tree and reach the goal. Breadth-first search (BFS) begins by expanding the top-level nodes of the search tree and then proceeds layer by layer into the depth. Depth-first search (DFS) recursively selects nodes from the search tree from shallow to deep and continues this process. Random search (RS) randomly selects an expandable node at each stage of expansion. Beam search (BS) selects $k$ nodes in each stage of expansion and can be viewed as a trade-off between BFS and RS.

We conducted experiments on 2 Intel i9-10900X, 1 AMD Ryzen 9 5900X, and 1 AMD Ryzen 9 7950X, running the search algorithms using multiple processes while maintaining a CPU utilization rate of 80\%. The maximum search depth was set to 15, and the beam size was set to 20. The total duration of the experiments was approximately 3 days. When the timeout for each problem was 300 seconds, the best success rate for problem-solving was approximately 30\%. When the timeout for each problem was increased to 600 seconds, the specific results are as follows.

\begin{table}
\centering
\caption{An overview of search results.}
\label{tab-search_results}
\begin{tabular}{ccccc}
\hline
\multirow{2}{*}{method} & \multirow{2}{*}{strategy} & \multicolumn{3}{c}{result (\%)} \\
\cline{3-5}
& & solved & unsolved & timeout \\
\hline
FW & BFS & 38.86 & \textbf{7.42} & 53.72 \\
FW & DFS & 36.16 & 9.80 & 54.05 \\
FW & RS & \textbf{39.71} & 9.07 & 51.22 \\
FW & BS & 25.28 & 38.72 & \textbf{36.00} \\
BW & BFS & \textbf{35.44} & 2.68 & 61.88 \\
BW & DFS & 33.73 & \textbf{2.42} & 63.84 \\
BW & RS & 34.05 & 2.65 & 63.30 \\
BW & BS & 34.39 & 12.86 & \textbf{52.74} \\
\hline
\end{tabular}
\end{table}

An overview of search-based automated problem-solving results is presented in Tab.~\ref{tab-search_results}. The highest problem-solving success rate was achieved by forward random search, reaching 39.708\%. Most of the remaining problems were due to timeouts. As timeout settings are extended and computational resources increase, the proportion of timeout problems is expected to decrease. The number of unsolved problems using beam search was significantly higher compared to other strategies. This is because when selecting $k$ branches, beam search occasionally discards the correct branch. Other contributing factors may include code bugs, equation solving timeouts, and the omission of theorems related to trigonometric.

In accordance with the length of the theorems required for problem-solving, we roughly categorize the difficulty of the questions into 6 levels, denoted as $l_1(length <= 2)$, $l_2 (3 <= length <= 4)$, $l_3 (5 <= length <= 6)$, $l_4 (7 <= length <= 8)$, $l_5 (9 <= length <= 10)$, $l_6 (length >= 11)$, with corresponding problem numbers of 2407, 1898, 1247, 824, 313 and 292. The success rates for solving geometric problems of varying difficulty are presented in Tab.~\ref{tab-success_rates}. As problem difficulty increases, the success rate of problem-solving rapidly declines. This phenomenon can be attributed to the fact that search-based problem-solving methods exhibit exponential growth in solving time as the length of the theorem sequence increases, often resulting in timeouts before achieving the goal. For problems of lower difficulty, backward search demonstrate a relatively higher success rate, while forward search outperforms in the case of more challenging problems.

\begin{table}
\centering
\caption{Results of success rates.}
\label{tab-success_rates}
\begin{tabular}{ccccccccc}
\hline
\multirow{2}{*}{method} & \multirow{2}{*}{strategy} & \multicolumn{7}{c}{success rates (\%)} \\
\cline{3-9}
& & total & $l_1$ & $l_2$ & $l_3$ & $l_4$ & $l_5$ & $l_6$ \\
\hline
FW & BFS & 38.86 & 59.95 & 38.62 & 28.55 & \textbf{17.35} & \textbf{8.63} & 3.77 \\
FW & DFS & 36.16 & 55.75 & \textbf{40.04} & 22.94 & 12.38 & 7.03 & 4.11 \\
FW & RS & \textbf{39.71} & 59.24 & \textbf{40.04} & \textbf{33.68} & 16.38 & 5.43 & \textbf{4.79} \\
FW & BS & 25.28 & 46.12 & 22.60 & 13.47 & 5.83 & 2.88 & 0.34 \\
BW & BFS & 35.44 & \textbf{67.22} & 33.72 & 11.15 & 6.67 & 6.07 & 1.03 \\
BW & DFS & 33.73 & 65.93 & 30.82 & 8.90 & 6.55 & 5.11 & 0.68 \\
BW & RS & 34.05 & 66.64 & 31.66 & 8.66 & 5.83 & 4.47 & 0.68 \\
BW & BS & 34.39 & 67.10 & 31.35 & 9.46 & 6.31 & 5.75 & 1.03 \\
\hline
\end{tabular}
\end{table}

\begin{figure}
\centering
\includegraphics[width=\textwidth]{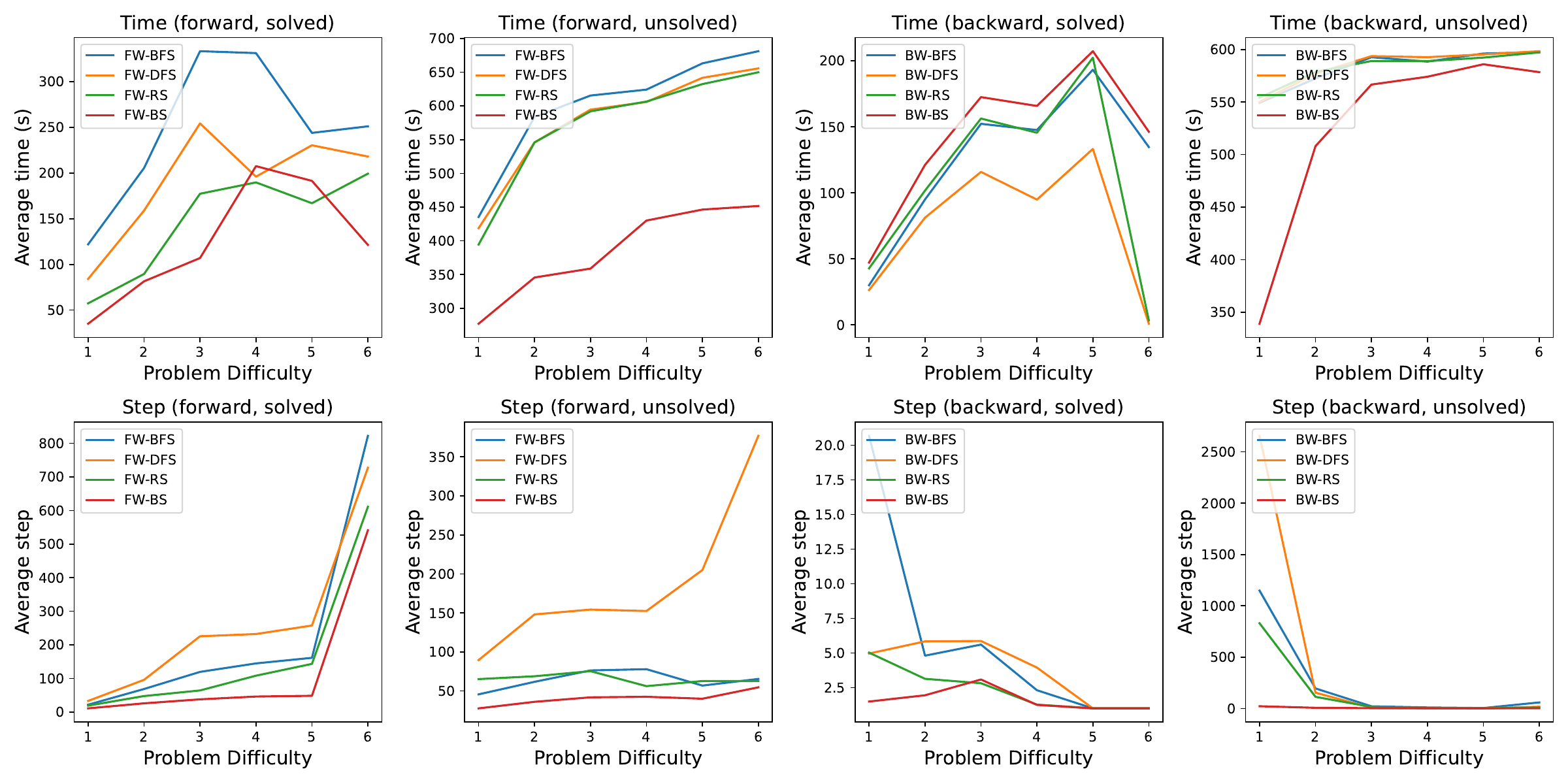}
\caption{Average search time (the top 4). Average search step (the bottom 4). The raw data for the charts can be found in App.~\ref{app-FGPS}}
\label{fig-chart_search_results}
\end{figure}

The efficiency of the problem-solving algorithm can be measured by the search time and step. The experimental results of search-based automated problem-solving algorithms on the formalgeo7k are presented in Fig.~\ref{fig-chart_search_results}.

In terms of average search time, backward search is slightly better than forward search overall. For solved problems, the search time is roughly proportional to the difficulty of the problems when problems are of low difficulty. However, as the difficulty increases, the search time for both forward search and backward search decreases. On the one hand, this is because there are very few successfully solved high-difficulty problems, leading to significant statistical errors. On the other hand, when dividing the difficulty of problems, we only consider the length of the solution theorem sequence but do not consider the time required for each theorem execution. The solved high-difficulty problems are precisely those that require less solution time. For unsolved problems, the search time is roughly proportional to the difficulty of the problems.

Comparing different search strategies, it can be observed that in the forward search, BFS has a slightly lower success rate compared to the RS, but it takes the most time. BS has the lowest success rate but the least time consumption. For forward search, RS is the optimal strategy as it has the highest success rate and only slightly higher time consumption than BS that has the lowest success rate. In backward search, BFS is the optimal strategy, with the highest success rate and only slightly higher time consumption than DFS.

We observe a significant difference in the solution time of the BS strategy in backward search for solved and unsolved problems. This difference may be due to the characteristics of the backward search, where even if possible solution branches were discarded in previous steps, they may be reconstructed in later search steps. Therefore, as for BS of backward search, discarding potential solusation branches does not lead to solution failure but takes longer search time.

Regarding the search step, forward search statistics are based on the number of nodes, while backward search statistics are based on the number of super nodes. Hence, they cannot be directly compared. The search step length in forward search is positively correlated with the difficulty of problems, while in backward search, it is negatively correlated with problem difficulty. The results of backward search are counterintuitive, and this could be because, for higher difficulty problems, the super nodes in backward search may contain more nodes, leading to increased time spent traversing a single super node and a reduction in the total number of traversed super nodes. Additionally, it can be observed that the search step length for unsolved problems in backward search is significantly higher than the average step length for solved problems. This is because, compared to forward search, backward search is less likely to halt, and it continues searching even if it misses a potential solution branch.

Comparing different strategies, DFS has the highest search step length, BS has the lowest search step length, and RS and BFS strategies have approximately the same average step length. For forward search, RS strategy is still the optimal strategy because it has the highest success rate and its search step length is only slightly higher than BS. Backward search does not exhibit a significantly superior strategy.

\section{Conclusion}

We have introduced GFT, which includes geometry ontology and geometry representation theory, to guide the formalization of geometric problems. Building upon GFT, we have developed the geometric formal system FormalGeo and constructed the solver FGPS. Furthermore, we have annotated the geometric problem-solving datasets, formalgeo7k and formalgeo-imo. Experiments have demonstrated the correctness and utility of GFT. We have also analyzed the success rate and efficiency of the solver's automatic problem-solving algorithm.

In the future, we plan to enhance GFT to make it more comprehensive and endowed with stronger representational capabilities. We also intend to further improve FormalGeo by expanding the types of predicates and theorems, as well as annotating formalgeo-imo datasets. Additionally, we aim to apply deep learning techniques to search tree pruning for the automatic solving of IMO-level geometric problems.

\section*{Acknowledgement}

This research was supported by NSFC Grant.12071282.

\section*{Contributions}

\textbf{Xiaokai Zhang:} Conceptualization, Methodology, Coding, Dataset annotation, Writing – original draft \& review \& editing. \textbf{Na Zhu:} Conceptualization, Methodology, Dataset annotation, Writing – review \& editing. \textbf{Yiming He, Jia Zou:} Conceptualization, Methodology, Dataset annotation. \textbf{Qike Huang, Xiaoxiao Jin, Yanjun Guo, Chenyang Mao, Zhe Zhu, Dengfeng Yue, Fangzhen Zhu, Yang Li, Yifan Wang, Yiwen Huang, Runan Wang, Cheng Qin:} Dataset annotation. \textbf{Zhenbing Zeng, Shaorong Xie, Xiangfeng Luo:} Writing – review. \textbf{Tuo Leng:} Supervision, Funding acquisition, Conceptualization, Methodology, Writing – review \& editing.

\bibliography{ref}

\appendix

\newpage

\section{Examples}

Guided by the GFT, we have developed the geometry formal system, FormalGeo. In this section, we provide several examples related to the design of FormalGeo predicates and theorems to help readers better understand the GFT. App.~\ref{app-GOD} and App.~\ref{app-GKG} correspond to the application of geometry ontology, while App.~\ref{app-PDL}, App.~\ref{app-TDL}, and App.~\ref{app-CDL} correspond to the application of geometry representation theory.

\subsection{Geometry ontology domain}
\label{app-GOD}

The geometry ontology domain comprehensively summarizes and categorizes geometric knowledge. It is divided into 4 quadrants along 2 dimensions: dynamic versus static, and number versus shape. Each quadrant encompasses mappings for axiomatic system, formal system, and solver, as depicted in Tab.~\ref{tab-god}.

\begin{table}
\centering
\caption{Examples of geometry ontology domain.}
\label{tab-god}
\begin{tabular}{ccp{8cm}}
\hline
\textbf{Quadrants} & \textbf{Tiers} & \textbf{Examples} \\
\hline
\multirow{3}{*}{\makecell{1 \\ Static \\ Shape}} & \makecell{axiomatic \\ system} & Triangle ABC is an equilateral triangle. \\
& \makecell{formal \\ system} & EquilateralTriangle(ABC) \\
& solver & $R_{ET} = \{ABC, BCA, CAB\}$ \\

\hline
\multirow{3}{*}{\makecell{2 \\ Static \\ Number}} & \makecell{axiomatic \\ system} & The length of line AB is equal to the length of line CD. \\
& \makecell{formal \\ system} & Equal(LengthOfLine(AB),LengthOfLine(CD)) \\
& solver & $ll_{AB} - ll_{CD} = 0$ \\

\hline
\multirow{3}{*}{\makecell{3 \\ Dynamic \\ Number}} & \makecell{axiomatic \\ system} & A triangle with two equal legs is an isosceles triangle. \\
& \makecell{formal \\ system} & \makecell[l]{Triangle(ABC)\& \\ Equal(LengthOfLine(AB),LengthOfLine(AC))$\rightarrow$ \\ IsoscelesTriangle(ABC)} \\
& solver & GPL defined in Eq.~\ref{eq-GPL3} \\

\hline
\multirow{3}{*}{\makecell{4 \\ Dynamic \\ Shape}} & \makecell{axiomatic \\ system} & If AB is parallel to CD, and CD is parallel to EF, then AB is parallel to EF. \\
& \makecell{formal \\ system} & Parallel(AB,CD)\&Parallel(CD,EF)$\rightarrow$Parallel(AB,EF) \\
& solver & GPL defined in Eq.~\ref{eq-GPL1} and Eq.~\ref{eq-GPL2} \\
\hline
\end{tabular}
\end{table}

\subsection{Geometry knowledge graph}
\label{app-GKG}

The geometry knowledge graph is an extension of the geometry ontology domain, which structurally presents the design process of predicate definition language and theorem definition language, preventing omissions or redundancies. FormalGeo comprises n predicates and m theorems, and its geometry knowledge graph is illustrated in Fig.~\ref{fig-geometry_knowledge_graph}.

\subsection{Predicate definition language}
\label{app-PDL}

FormalGeo comprises 88 predicates, including 25 fundamental predicates (Tab.~\ref{tab-pdl1}) built into the solver and 12 entities (Tab.~\ref{tab-pdl2}), 30 entity relationships (Tab.~\ref{tab-pdl3}), and 21 attributions (Tab.~\ref{tab-pdl4}) defined using the predicate definition language. The structured relationships between predicates are depicted in Fig.~\ref{fig-geometry_knowledge_graph}.

The detailed statements for defining a predicate is as shown in the Tab.~\ref{tab-pdl}, including the predicate name and point variable declaration, validity check declaration, multiple representations, and automatic expansion. Additionally, when defining attributes, it also includes symbolic form declaration.

\begin{table}
\centering
\caption{Detailed statement examples for defining a predicate.}
\label{tab-pdl}
\begin{tabular}{ccl}
\hline
name & item & content\\
\hline
\multirow{4}{*}{IsMidpointOfLine(M,AB)} & ee\_check & \makecell[l]{Point(M) \\ Line(AB) \\ Collinear(AMB)} \\
& fv\_check & M,AB \\
& multi & M,BA \\
& extend &Equal(LengthOfLine(AM),LengthOfLine(MB)) \\
\hline
\multirow{3}{*}{LengthOfLine(AB)} & ee\_check & Line(AB) \\
& sym & ll \\
& multi & BA \\
\hline
\end{tabular}
\end{table}

\subsection{Theorem definition language}
\label{app-TDL}

\begin{figure}
\centering
\includegraphics[width=\textwidth]{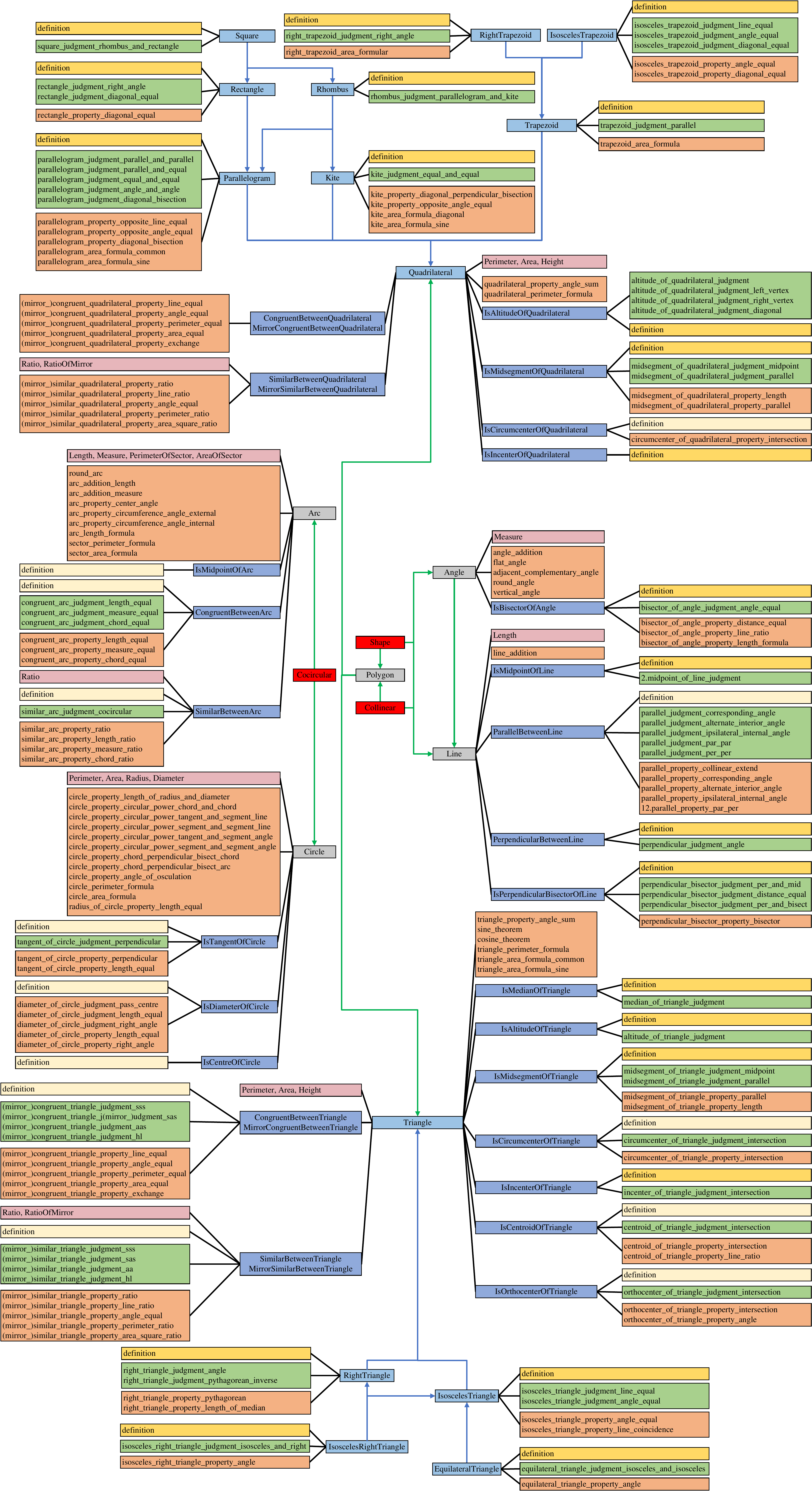}
\caption{Geometry knowledge graph of FormalGeo}
\label{fig-geometry_knowledge_graph}
\end{figure}

Theorems are defined using the GPL, comprising two parts: premises and conclusions, as shown in the Tab.~\ref{tab-tdl}. FormalGeo encompasses 196 theorems, and their structured relationships are illustrated in Fig.~\ref{fig-geometry_knowledge_graph}.

\begin{table}
\centering
\caption{Detailed statement examples for defining a theorem.}
\label{tab-tdl}
\begin{tabular}{ccl}
\hline
name & item & content\\
\hline
\multirow{2}{*}{\makecell{midpoint\_of\_line\_\\judgment(M,AB)}} & premise & \makecell[l]{Collinear(AMB)\&\\Equal(LengthOfLine(AM),LengthOfLine(MB))} \\
& conclusion & IsMidpointOfLine(M,AB) \\
\hline
\multirow{2}{*}{\makecell{vertical\_angle\\(AOC,BOD)}} & premise & \makecell[l]{Collinear(AOB)\&Collinear(COD)\&\\Angle(AOC)\&Angle(BOD)} \\
& conclusion & Equal(MeasureOfAngle(AOC),MeasureOfAngle(BOD)) \\
\hline
\end{tabular}
\end{table}

\subsection{Condition declaration language}
\label{app-CDL}

\begin{figure}
\centering
\includegraphics[width=\textwidth]{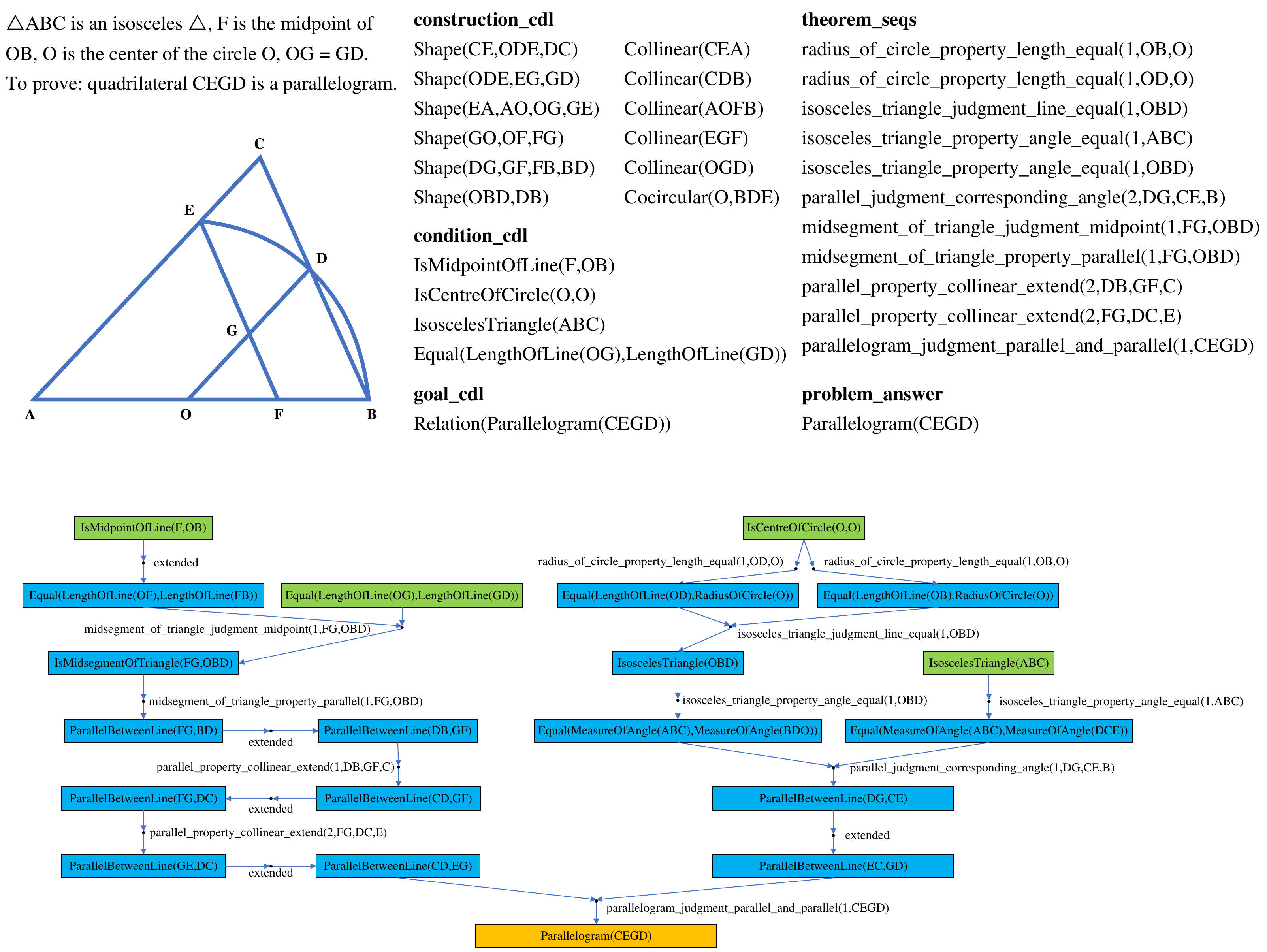}
\caption{An example of annotated problem and solution hypertree. For layout considerations, hypertree in this image omits many details. The original hypertree generated by the solver during the problem-solving process can be seen in Fig.~\ref{fig-hyper_tree_origin}.}
\label{fig-problem_example}
\end{figure}

We can use CDL to transform the description of geometric problems into a formal language. CDL consists of three main parts: construction statements, condition statements, and goal statements. An example of a transformed problem is illustrated in Fig.~\ref{fig-problem_example}, with theorem\_seqs denoting the annotated problem-solving theorem sequence.

We abstract the process of geometry problem-solving as a hyper tree, where tree root nodes represent known conditions (labeled in green in Fig.~\ref{fig-problem_example}), tree hyperedges represent geometric theorems, and the problem-solving process is a path from the root node to the target leaf node (labeled in yellow in Fig.~\ref{fig-problem_example}). 

\begin{figure}
\centering
\includegraphics[width=\textwidth]{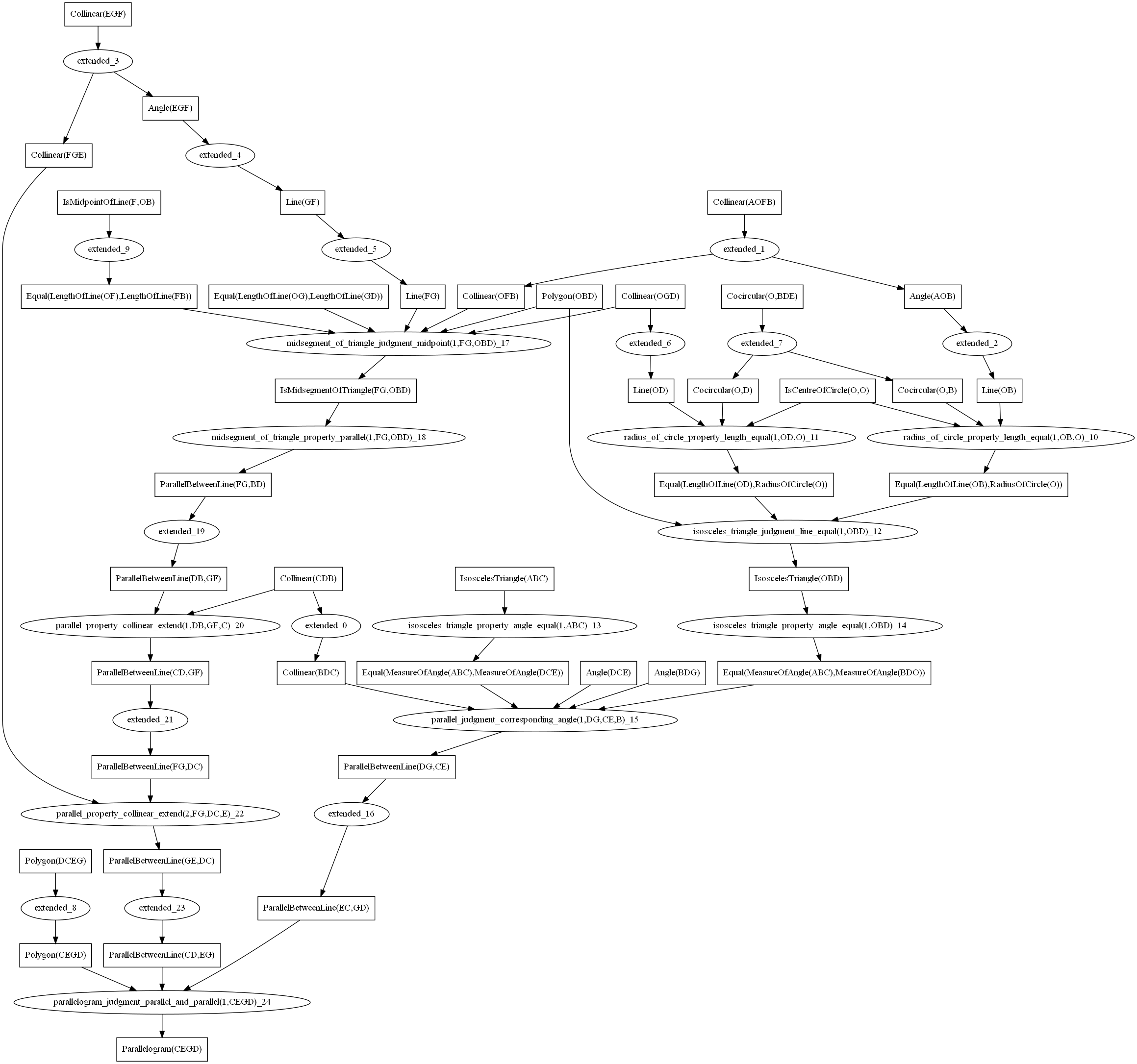}
\caption{The original hypertree}
\label{fig-hyper_tree_origin}
\end{figure}

\begin{table}
\centering
\caption{Predicates built into the solver.}
\label{tab-pdl1}
\begin{tabular}{cccc}
\hline
id & type & name & examples \\
\hline
1 & Construction & Shape & Shape(AB,BC,CA) \\
2 & Construction & Collinear & Collinear(ABCD) \\
3 & Construction & Cocircular & Cocircular(O,ABC) \\
4 & BasicEntity & Point & Point(A) \\
5 & BasicEntity & Line & Line(AB) \\
6 & BasicEntity & Arc & Arc(OAB) \\
7 & BasicEntity & Angle & Angle(ABC) \\
8 & BasicEntity & Polygon & Polygon(ABCD) \\
9 & BasicEntity & Circle & Circle(O) \\
10 & Algebra & Equal & Equal(a,b) \\
11 & Algebra & Equation & Equation(a-b) \\
12 & Attribution & Free & Free(x) \\
13 & Operation & Add & Equal(Add(a,b,c),1) \\
14 & Operation & Sub & Equal(Sub(a,b),1) \\
15 & Operation & Mul & Equal(Mul(a,b,c),1) \\
16 & Operation & Div & Equal(Div(a,b),1) \\
17 & Operation & Pow & Equal(Pow(a,b),1) \\
18 & Operation & Mod & Equal(Mod(a,b),1) \\
19 & Operation & Sqrt & Equal(Sqrt(a),1) \\
20 & Operation & Sin & Equal(Sin(a),1/2) \\
21 & Operation & Cos & Equal(Cos(a),1/2) \\
22 & Operation & Tan & Equal(Tan(a),1) \\
23 & Target & Value & Value(a) \\
24 & Target & Equal & Equal(a,b) \\
25 & Target & Relation & Relation(RightTriangle(ABC)) \\
\hline
\end{tabular}
\end{table}

\begin{table}
\centering
\caption{Entities defined using the predicate definition language.}
\label{tab-pdl2}
\begin{tabular}{cccc}
\hline
id & type & examples \\
\hline
26 & Entity & RightTriangle(ABC) \\ 
27 & Entity & IsoscelesTriangle(ABC) \\ 
28 & Entity & IsoscelesRightTriangle(ABC) \\ 
29 & Entity & EquilateralTriangle(ABC) \\ 
30 & Entity & Kite(ABCD) \\ 
31 & Entity & Parallelogram(ABCD) \\ 
32 & Entity & Rhombus(ABCD) \\ 
33 & Entity & Rectangle(ABCD) \\ 
34 & Entity & Square(ABCD) \\ 
35 & Entity & Trapezoid(ABCD) \\ 
36 & Entity & IsoscelesTrapezoid(ABCD) \\ 
37 & Entity & RightTrapezoid(ABCD) \\ 
\hline
\end{tabular}
\end{table}

\begin{table}
\centering
\caption{Relations defined using the predicate definition language.}
\label{tab-pdl3}
\begin{tabular}{cccc}
\hline
id & type & examples \\
\hline
38 & Relation & IsMidpointOfLine(M,AB) \\ 
39 & Relation & IsMidpointOfArc(M,OAB) \\ 
40 & Relation & ParallelBetweenLine(AB,CD) \\ 
41 & Relation & PerpendicularBetweenLine(AC,BC) \\ 
42 & Relation & IsPerpendicularBisectorOfLine(AB,CD) \\ 
43 & Relation & IsBisectorOfAngle(BD,ABC) \\ 
44 & Relation & IsMedianOfTriangle(AD,ABC) \\ 
45 & Relation & IsAltitudeOfTriangle(AD,ABC) \\ 
46 & Relation & IsMidsegmentOfTriangle(DE,ABC) \\ 
47 & Relation & IsCircumcenterOfTriangle(O,ABC) \\ 
48 & Relation & IsIncenterOfTriangle(O,ABC) \\ 
49 & Relation & IsCentroidOfTriangle(O,ABC) \\ 
50 & Relation & IsOrthocenterOfTriangle(O,ABC) \\ 
51 & Relation & CongruentBetweenTriangle(ABC,DEF) \\ 
52 & Relation & MirrorCongruentBetweenTriangle(ABC,DEF) \\ 
53 & Relation & SimilarBetweenTriangle(ABC,DEF) \\ 
54 & Relation & MirrorSimilarBetweenTriangle(ABC,DEF) \\ 
55 & Relation & IsAltitudeOfQuadrilateral(EF,ABCD) \\ 
56 & Relation & IsMidsegmentOfQuadrilateral(EF,ABCD) \\ 
57 & Relation & IsCircumcenterOfQuadrilateral(O,ABCD) \\ 
58 & Relation & IsIncenterOfQuadrilateral(O,ABCD) \\ 
59 & Relation & CongruentBetweenQuadrilateral(ABCD,EFGH) \\ 
60 & Relation & MirrorCongruentBetweenQuadrilateral(ABCD,EFGH) \\ 
61 & Relation & SimilarBetweenQuadrilateral(ABCD,EFGH) \\ 
62 & Relation & MirrorSimilarBetweenQuadrilateral(ABCD,EFGH) \\ 
63 & Relation & CongruentBetweenArc(OAB,OCD) \\ 
64 & Relation & SimilarBetweenArc(OAB,OCD) \\ 
65 & Relation & IsDiameterOfCircle(AB,O) \\ 
66 & Relation & IsTangentOfCircle(PA,O) \\ 
67 & Relation & IsCentreOfCircle(P,O) \\
\hline
\end{tabular}
\end{table}

\begin{table}
\centering
\caption{Attributions defined using the predicate definition language.}
\label{tab-pdl4}
\begin{tabular}{cccc}
\hline
id & type & examples \\
\hline
68 & Attribution & LengthOfLine(AB) \\ 
69 & Attribution & MeasureOfAngle(ABC) \\ 
70 & Attribution & PerimeterOfTriangle(ABC) \\ 
71 & Attribution & AreaOfTriangle(ABC) \\ 
72 & Attribution & HeightOfTriangle(ABC) \\ 
73 & Attribution & RatioOfSimilarTriangle(ABC) \\ 
74 & Attribution & RatioOfMirrorSimilarTriangle(ABC) \\ 
75 & Attribution & PerimeterOfQuadrilateral(ABCD) \\ 
76 & Attribution & AreaOfQuadrilateral(ABCD) \\ 
77 & Attribution & HeightOfQuadrilateral(ABCD) \\ 
78 & Attribution & RatioOfSimilarQuadrilateral(ABCD) \\ 
79 & Attribution & RatioOfMirrorSimilarQuadrilateral(ABCD) \\ 
80 & Attribution & LengthOfArc(OAB) \\ 
81 & Attribution & MeasureOfArc(OAB) \\ 
82 & Attribution & RatioOfSimilarArc(OAB) \\ 
83 & Attribution & RadiusOfCircle(O) \\ 
84 & Attribution & DiameterOfCircle(O) \\ 
85 & Attribution & PerimeterOfCircle(O) \\ 
86 & Attribution & AreaOfCircle(O) \\ 
87 & Attribution & PerimeterOfSector(OAB) \\ 
88 & Attribution & AreaOfSector(OAB) \\ 
\hline
\end{tabular}
\end{table}

\newpage

\section{Consistency theory of formal system}
\label{app-consistency_theory}

A formal system is an abstraction of the real world that simplifies and clarifies problems or concepts by removing certain details and complexities. Its primary objective is to provide a precise, consistent, and reliable method for studying and solving problems while eliminating ambiguity and uncertainty. By formalizing problems, we can apply mathematical, logical, and computational methods to address a wide range of real-world issues.

In the real world, a concept $c$, which could represent anything, the properties of things, relationships between things, etc., can be represented by a symbol $s^{(c)}$ within the formal system. A single concept may have multiple symbol representations within the formal system, denoted as the set $R_c$. The rules governing the conversion of concepts in the real world are represented as various rules in formal system, denoted as $\xrightarrow{r}$. These rules define relationships and operations between symbols, determining the methods of inference and computation.

When designing a formal system, it is essential to ensure the consistency of static representations. For any symbol representation $s^{(c)}_i$ of a concept $c$, there should exist an operation $multi$ to obtain the symbol set $R_c$, as shown in Eq.~\ref{eq-ct1}. The mapping between $c$ and $R_c$ should be a reversible mapping, as indicated in Eq.~\ref{eq-ct2}.

\begin{equation}
multi(s^{(c)}_i) = R_c, s^{(c)}_i \in R_c
\label{eq-ct1}
\end{equation}

\begin{equation}
f(c) = R_c, f^{-1}(R_c) = c
\label{eq-ct2}
\end{equation}

Additionally, the design of a formal system should ensure the consistency of dynamic processes. For any rule $\xrightarrow{r}$ designed to govern the conversion of concepts $c_a$ and $c_b$ in the real world, it should satisfy the condition that, for any symbol representation $s^{(c_a)}_i$ of $c_a$, the application of $\xrightarrow{r}$ results in and can only yield the symbol representation $s^{(c_b)}_j$ of $c_b$.

\begin{equation}
c_a \xrightarrow{r} c_b
\label{eq-ct3}
\end{equation}

\begin{equation}
s^{(c_a)}_i \xrightarrow{r} s^{(c_b)}_j, s^{(c_a)}_i \in R_{c_a}, s^{(c_b)}_j \in R_{c_b}
\label{eq-ct4}
\end{equation}

\section{Topological mapping method}
\label{app-TMM}
When we utilize topological mapping method to transform the TSI of a diagram into a formal representation, we can define the fundamental transformations of the diagram as operations on its TSI formal representation.

The formal representation of the TSI of a geometric diagram is denoted as $(p_1, p_2, \ldots, p_n)$. Clockwise rotation of the diagram can be defined as moving the first point in the sequence to the end of the sequence, as shown in Eq.~\ref{eq-rotate}. Reflection can be defined as reversing the order of the sequence, as shown in Eq.~\ref{eq-reflect}. Translation, shearing, and scaling do not alter the TSI of the diagram, as shown in Eq.~\ref{eq-linear}.

\begin{equation}
rotate((p_1, p_2, \dots, p_n)) = (p_2, \dots, p_n, p_1)
\label{eq-rotate}
\end{equation}

\begin{equation}
reflect((p_1, p_2, \dots, p_n)) = (p_n, p_{n-1}, \dots, p_1)
\label{eq-reflect}
\end{equation}

\begin{equation}
linear((p_1, p_2, \dots, p_n)) = (p_1, p_2, \dots, p_n)
\label{eq-linear}
\end{equation}

\section{Topological construction method}
\label{app-TCM}
The operation $\oplus$ defined by Eq.~\ref{eq-TCM1} and Eq.~\ref{eq-TCM2} is meaningful only when two shapes are non-overlapping simple closed shapes and when the two shapes share adjacent edges (the number of common points $>$ 2).

The formal representation of composite diagrams, along with $\oplus$, forms a semigroup, satisfying closure, commutative law and associative law. The formal representation of composite diagrams is defined as the collection of the formal representations of all its constituent shapes, so closure is evidently satisfied. Next, we will prove that the operation $\oplus$ satisfies commutative law and associative law.

If diagram $C$ is composed of diagrams $A$ and $B$, their formal representations of TSI are represented as sets $R_a$, $R_b$, and $R_c$, respectively. Diagram $A$, $B$, $C$ contains $m$, $n$, $o$ points, respectively. There are $x$ ($x\geq2$)common points shared between $A$ and $B$ and $y$ ($y\geq2$)common points shared between $B$ and $C$. The formal representations of their TSI can be represented as Eq.~\ref{eq-app_tcm1}, Eq.~\ref{eq-app_tcm2} and Eq.~\ref{eq-app_tcm3}.

\begin{equation}
\begin{aligned}
    P_a=(p^{(a)}_1, \dots, p^{(a)}_{s_1}, 
        p^{(a,b)}_{i}, \dots, p^{(a,b)}_{i+x}, 
        p^{(a)}_{s_2}, \dots, p^{(a)}_l)
\end{aligned}
\label{eq-app_tcm1}
\end{equation}

\begin{equation}
\begin{aligned}
    P_b=&(p^{(b)}_1, \dots, p^{(b)}_{s_1}, 
        p^{(a,b)}_{i+x}, \dots, p^{(a,b)}_{i},\\ &
        p^{(b)}_{s_2}, \dots, p^{(b)}_{s_3},
        p^{(b,c)}_{j}, \dots, p^{(b,c)}_{j+y},
        p^{(b)}_{s_4}, \dots, p^{(b)}_m)
\end{aligned}
\label{eq-app_tcm2}
\end{equation}

\begin{equation}
\begin{aligned}
    P_c=(p^{(c)}_1, \dots, p^{(c)}_{s_1}, 
        p^{(b,c)}_{j+y}, \dots, p^{(b,c)}_{j}, 
        p^{(c)}_{s_2}, \dots, p^{(c)}_n)
\end{aligned}
\label{eq-app_tcm3}
\end{equation}

We shall begin by proving the commutative law. As demonstrated in Eq.~\ref{eq-app_tcm4} and Eq.~\ref{eq-app_tcm5}, $P_a \otimes P_b = rotate^{(i)}(P_b \otimes P_a)$, where $i=|\{p^{(b)}_1, \dots, p^{(b)}_{s_1}, p^{(a,b)}_{i+x}, p^{(a)}_{s_2}, \dots, p^{(a)}_l\}|$. Then the commutative law can be demonstrated by Eq.~\ref{eq-app_tcm6}.

\begin{equation}
\begin{aligned}
    &P_a \otimes P_b \\
    =&(p^{(a)}_1, \dots, p^{(a)}_{s_1}, p^{(a,b)}_{i},
       p^{(b)}_{s_2}, \dots, p^{(b)}_{s_3},
       p^{(b,c)}_{j}, \dots,\\ &p^{(b,c)}_{j+y},
       p^{(b)}_{s_4}, \dots, p^{(b)}_m,
       p^{(b)}_1, \dots, p^{(b)}_{s_1}, p^{(a,b)}_{i+x},
       p^{(a)}_{s_2}, \dots, p^{(a)}_l)
\end{aligned}
\label{eq-app_tcm4}
\end{equation}

\begin{equation}
\begin{aligned}
    &P_b \otimes P_a \\
    =&(p^{(b)}_1, \dots, p^{(b)}_{s_1}, p^{(a,b)}_{i+x},
       p^{(a)}_{s_2}, \dots, p^{(a)}_l,
       p^{(a)}_1, \dots,\\ &p^{(a)}_{s_1}, p^{(a,b)}_{i},
       p^{(b)}_{s_2}, \dots, p^{(b)}_{s_3},
       p^{(b,c)}_{j}, \dots, p^{(b,c)}_{j+y},
       p^{(b)}_{s_4}, \dots, p^{(b)}_m)
\end{aligned}
\label{eq-app_tcm5}
\end{equation}

\begin{equation}
\begin{aligned}
    &R_a \oplus R_b\\
    =&\{rotate^{(i)}(P_a \otimes P_b)|i=1,2, \dots, l+m-2x+2\}\\
    =&\{rotate^{(i)}(P_b \otimes P_a)|i=1,2, \dots, l+m-2x+2\}\\
    =&R_b \oplus R_a
\end{aligned}
\label{eq-app_tcm6}
\end{equation}

Next, we proceed to prove the associative law. As demonstrated in Eq.~\ref{eq-app_tcm7} and Eq.~\ref{eq-app_tcm8}, $(P_a \otimes P_b) \otimes P_c = P_a \otimes (P_b \otimes P_c)$, where Then the commutative law can be demonstrated by Eq.~\ref{eq-app_tcm9}.

\begin{equation}
\begin{aligned}
    &(P_a \otimes P_b) \otimes P_c \\
    =&(p^{(a)}_1, \dots, p^{(a)}_{s_1}, p^{(a,b)}_{i},
       p^{(b)}_{s_2}, \dots, p^{(b)}_{s_3},
       p^{(b,c)}_{j}, \dots,\\ &p^{(b,c)}_{j+y},
       p^{(b)}_{s_4}, \dots, p^{(b)}_m,
       p^{(b)}_1, \dots, p^{(b)}_{s_1}, p^{(a,b)}_{i+x},
       p^{(a)}_{s_2}, \dots, p^{(a)}_l) 
       \\ &\otimes (p^{(c)}_1, \dots, p^{(c)}_{s_1}, 
       p^{(b,c)}_{j+y}, \dots, p^{(b,c)}_{j}, 
       p^{(c)}_{s_2}, \dots, p^{(c)}_n)\\
    =&(p^{(a)}_1, \dots, p^{(a)}_{s_1}, p^{(a,b)}_{i},
       p^{(b)}_{s_2}, \dots, p^{(b)}_{s_3},
       p^{(b,c)}_{j}, p^{(c)}_{s_2}, \dots, p^{(c)}_n,
       p^{(c)}_1, \dots, \\ &p^{(c)}_{s_1}, p^{(b,c)}_{j+y},
       p^{(b)}_{s_4}, \dots, p^{(b)}_m,
       p^{(b)}_1, \dots, p^{(b)}_{s_1}, p^{(a,b)}_{i+x},
       p^{(a)}_{s_2}, \dots, p^{(a)}_l)
\end{aligned}
\label{eq-app_tcm7}
\end{equation}

\begin{equation}
\begin{aligned}
    &P_a \otimes (P_b \otimes P_c) \\
    =&(p^{(a)}_1, \dots, p^{(a)}_{s_1}, 
       p^{(a,b)}_{i}, \dots, p^{(a,b)}_{i+x}, 
       p^{(a)}_{s_2}, \dots, p^{(a)}_l) \\
       & \otimes (p^{(b)}_1, \dots, p^{(b)}_{s_1}, 
       p^{(a,b)}_{i+x}, \dots, p^{(a,b)}_{i},
       p^{(b)}_{s_2}, \dots, p^{(b)}_{s_3},
       p^{(b,c)}_{j},p^{(c)}_{s_2}, \dots, \\ 
       &p^{(c)}_n, p^{(c)}_1, \dots, p^{(c)}_{s_1}, 
       p^{(b,c)}_{j+y}, p^{(b)}_{s_4}, \dots, p^{(b)}_m)\\
    =&(p^{(a)}_1, \dots, p^{(a)}_{s_1}, 
       p^{(a,b)}_{i},p^{(b)}_{s_2}, \dots, p^{(b)}_{s_3},
       p^{(b,c)}_{j},p^{(c)}_{s_2}, \dots,
       p^{(c)}_n, p^{(c)}_1, \dots,  \\ 
       &p^{(c)}_{s_1}, p^{(b,c)}_{j+y}, p^{(b)}_{s_4}, \dots, p^{(b)}_m,
       p^{(b)}_1, \dots, p^{(b)}_{s_1}, 
       p^{(a,b)}_{i+x}, p^{(a)}_{s_2}, \dots, p^{(a)}_l)  
\end{aligned}
\label{eq-app_tcm8}
\end{equation}

\begin{equation}
\begin{aligned}
    &(R_a \oplus R_b) \oplus R_c \\
    =&\{rotate^{(i)}((P_a \otimes P_b) \otimes P_c)|i=1,2, \dots, l+m+n-2x-2y+4\} \\
    =&\{rotate^{(i)}(P_a \otimes (P_b \otimes P_c))|i=1,2, \dots, l+m+n-2x-2y+4\} \\
    =&R_a \oplus (R_b \oplus R_c)
\end{aligned}
\label{eq-app_tcm9}
\end{equation}

The aforementioned properties enable us to construct diagrams from any two simple components of a composite diagram without considering the order of construction. This laws makes the TCM an order-independent diagram construction method. When formalizing problems and implementing TCM, we no longer need to be concerned about the order of construction statements.

We have implemented TCM in FGPS, and the algorithm description can be found in Alg.~\ref{alg-TCM}, where the $multi$ function extends a single TSI formal representation of a diagram to the set of all formal representations, as defined in Eq.~\ref{eq-TCM2}, and the $combine$ function is used to achieve the combination representation of two simple diagrams, as defined in Eq.~\ref{eq-TCM1}.

The computational cost of the algorithm per iteration is shown in Tab.~\ref{tab-tcm_complexity}, where $n$ represents the number of simple closed diagrams that make up the complex diagram. For the complex diagram, it takes at most $n-1$ combinations of simple figures to obtain the complex diagram. The number of combined diagrams generated after each iteration depends on the number of edges in the simple diagrams. For example, when $n$ triangles are combined, the number of quadrilaterals obtained is approximately $3n/2$, denoted as $k_i \cdot n$.

\begin{table}
\centering
\caption{Time Complexity Analysis of the implemented TCM.}
\label{tab-tcm_complexity}
\begin{tabular}{ccccc}
\hline
iteration & len(Units) & len(Combs) & combination times & len(NewCombs) \\
\hline
1 & $n$ & $n$ & $n \cdot n$ & $k_1 \cdot n$ \\
2 & $n$ & $k_1 \cdot n$ & $n \cdot k_1 \cdot n$ & $k_2 \cdot n$ \\
$\dots$ & $\dots$ & $\dots$ & $\dots$ & $\dots$ \\
$n-1$ & $n$ & $k_{n-2} \cdot n$ & $n \cdot k_{n-2} \cdot n$ & 0 \\
\hline
\end{tabular}
\end{table}

The time complexity of the implemented TCM is given by Eq.~\ref{eq-tcm_complexity}.

\begin{equation}
\begin{aligned}
    n^2 + k_1n^2 + \dots + k_{n-2}n^2
    =(1 + k_1 + \dots + k_{n-2})n^2
    \approx O(n^3)
\end{aligned}
\label{eq-tcm_complexity}
\end{equation}

\begin{algorithm}
\caption{Topological construction method}
\label{alg-TCM}
\begin{algorithmic}
\STATE \textbf{Input:} $Units$: list, TSI of simple diagrams that make up combined diagram.
\STATE \textbf{Output:} $Results$: list, constructed TSI of combined diagram.
\STATE $Units \leftarrow multi(Units)$
\STATE $Combs \leftarrow Units$
\STATE $Results \leftarrow Combs$
\WHILE{$len(Combs) > 0$}
    \STATE Initialize an list $NewCombs$
    \FOR{$i = 1$ to $len(Units)$}
        \STATE $unit \leftarrow Units[i]$ 
        \FOR{$j = 1$ to $len(Combs)$} 
            \STATE $comb \leftarrow Combs[j]$ 
            \STATE $new\_comb \leftarrow combine(unit,comb)$
            \IF{$new\_comb$ is not $None$ \AND $new\_comb$ is not in $Results$ }
                \STATE $new\_combs \leftarrow multi(new\_comb)$
                \FOR{$k = 1$ to $len(new\_combs)$}
                    \STATE add $new\_combs[k]$ to $NewCombs$
                    \STATE add $new\_combs[k]$ to $Results$
                \ENDFOR
            \ENDIF
        \ENDFOR
    \ENDFOR
    \STATE $Combs \leftarrow NewCombs$
\ENDWHILE
\end{algorithmic}
\end{algorithm}

\section{Geometry predicate logic}
\label{app-GPL}

The GPL operation $\&$ is, in fact, a more general form of operation $\oplus$. $\oplus$ requires that the two geometric relations involved in the operation are formalized TSI, and the result has a fixed form. $\&$ is not limited to the same type of geometric relations and can even operate across different types of relations, such as geometric and quantitative relations. The result of the operation is a set of point variables, which can be combined into specific structures as needed. In addition, GPL has introduced the $|$ and $\sim$ to enhance its expressive power.

For the $\&$ operation, when the second relation is a quantitative relation, it is denoted as $R_1 \& R_A$. The operation is only meaningful when the point variables of $R_A$ are a subset of those in $R_1$. As for the $|$ operation, it is only meaningful when the two relations involved have the same point variable structure. The $\&$ operation satisfies commutative law and associative law, while the combined operation of $\&$ and $|$ satisfies the distributive law.

\begin{equation}
R_1(v^{(1)}_{1}, v^{(1)}_{2}, \dots, v^{(1)}_{l})
\label{eq-app_gpl1}
\end{equation}
\begin{equation}
R_2(v^{(2)}_{1}, v^{(2)}_{2}, \dots, v^{(2)}_{m})
\label{eq-app_gpl2}
\end{equation}
\begin{equation}
R_3(v^{(3)}_{1}, v^{(3)}_{2}, \dots, v^{(3)}_{n})
\label{eq-app_gpl3}
\end{equation}

There are 3 relations as shown in Eq.~\ref{eq-app_gpl1}, Eq.~\ref{eq-app_gpl2} and Eq.~\ref{eq-app_gpl3}. In the subsequent proof process, for the sake of clarity, we temporarily omit the operation of removing duplicate variables, as defined in Eq.~\ref{eq-GPL2}. 

\begin{equation}
\begin{aligned}
    &R_1 \& R_2 \\
    =&\{(r^{1}_i, r^{2}_j) | (r^{1}_i, r^{2}_j) \in R_1 \times R_2, r^{1}_i(v) = r^{2}_j(v), v \in V_1 \cap V_2 \}\\
    =&\{(r^{2}_j, r^{1}_i) | (r^{2}_j, r^{1}_i) \in R_2 \times R_1, r^{1}_i(v) = r^{2}_j(v), v \in V_1 \cap V_2 \}\\
    =&R_2 \& R_1
\end{aligned}
\label{eq-app_gpl4}
\end{equation}

\begin{equation}
\begin{aligned}
    &(R_1 \& R_2) \& R_3\\
    =&\{(r^{1}_i, r^{2}_j, r^{3}_k) | (r^{1}_i, r^{2}_j, r^{3}_k) \in (R_1 \times R_2) \times R_3, r^{1}_i(v) = r^{2}_j(v), v \in V_1 \cap V_2,\\&r^{1}_i(u) = r^{3}_k(u), u \in V_1 \cap V_3, r^{2}_j(w) = r^{3}_k(w), w \in V_2 \cap V_3 \}\\
    =&\{(r^{1}_i, r^{2}_j, r^{3}_k) | (r^{1}_i, r^{2}_j, r^{3}_k) \in R_1 \times (R_2 \times R_3), r^{1}_i(v) = r^{2}_j(v), v \in V_1 \cap V_2,\\&r^{1}_i(u) = r^{3}_k(u), u \in V_1 \cap V_3, r^{2}_j(w) = r^{3}_k(w), w \in V_2 \cap V_3 \}\\
    =&R_1 \& (R_2 \& R_3)
\end{aligned}
\label{eq-app_gpl5}
\end{equation}

\begin{equation}
\begin{aligned}
    &R_1 \& (R_2 | R_3)\\
    =&\{(r^{1}_i, r^{2}_j) | (r^{1}_i, r^{2}_j) \in R_1 \times (R_2 \cup R_3), r^{1}_i(v) = r^{2}_j(v), v \in V_1 \cap (V_2 \cup V_3) \}\\
    =&\{(r^{1}_i, r^{2}_j) | (r^{1}_i, r^{2}_j) \in R_1 \times R_2, r^{1}_i(v) = 
        r^{2}_j(v), v \in V_1 \cap V_2 \} \\
      &\cup \{(r^{1}_i, r^{2}_j) | (r^{1}_i, r^{2}_j) \in R_1 \times R_3, r^{1}_i(v) = 
              r^{2}_j(v), v \in V_1 \cap V_3 \}\\
    =&(R_1 \& R_2) | (R_1 \& R_3)
\end{aligned}
\label{eq-app_gpl6}
\end{equation}

The proof of ommutative law, associative law and distributive law can be found in Eq.~\ref{eq-app_gpl4}, Eq.~\ref{eq-app_gpl5} and Eq.~\ref{eq-app_gpl6}. We utilize the laws of the Cartesian product $\times$ to prove the laws of the $\&$. The elements of the relation $R$ are treated as unordered sets. Therefore, $(r^{1}_i, r^{2}_j)$ and $(r^{2}_j, r^{1}_i)$ are considered equivalent in Eq.~\ref{eq-app_gpl4}. As mentioned earlier, $|$ is meaningful only when the two relations involved have the same point variable structure. Hence, in Eq.~\ref{eq-app_gpl6}, we have $V_2 = V_3 = V_2 \cup V_3$. The aforementioned laws allow us to simplify GPL statements before their execution, speeding up the process.

\section{FGPS}
\label{app-FGPS}

\begin{figure}
\centering
\includegraphics[width=\textwidth]{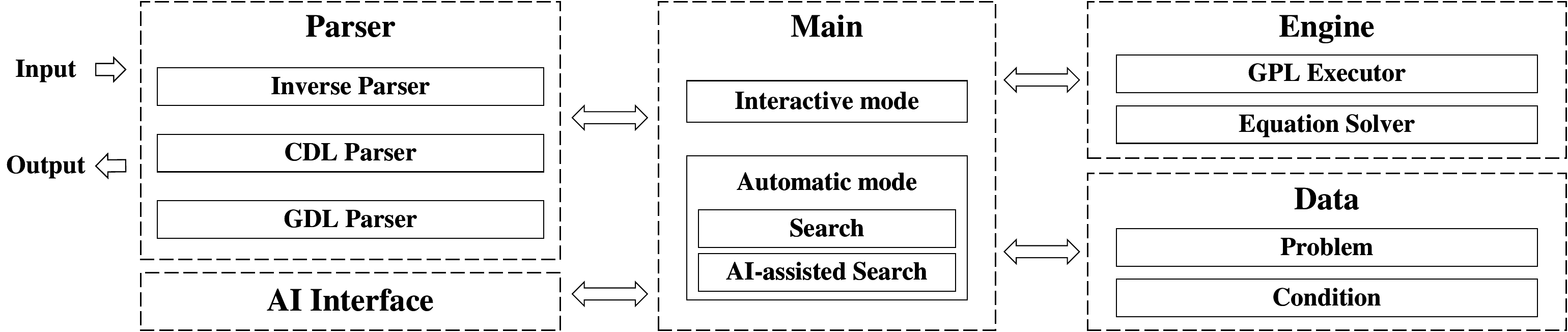}
\caption{The architecture of FGPS}
\label{fig-solver_architecture}
\end{figure}

The structure of FGPS can be divided into five main components: Main Logic Control, Core Solving Engine, Formal Language Parser, Data Loader, and AI Interface. The relationships between these modules are illustrated in Fig.~\ref{fig-solver_architecture}.

\textbf{Main} is control module of FGPS, invoking other modules to enable interactive problem-solving and automated problem-solving. The automated solving component implements both forward search and backward search, allowing for the configuration of various search strategies (breadth-first, depth-first, random, beam) and defining interfaces for AI-assisted searches.

\textbf{Engine} is the core component of FGPS, responsible for parsing and executing GPL and consists of two sub-modules, GPL Executor for relational inference and Equation Solver for algebraic computation.

\textbf{Parser} facilitates bidirectional conversion between formal language and machine language. It consists of 3 sub-modules. GDL Parser parses GPDL and GTDL into machine language, enabling custom configuration of the Solver. CDL Parser parses the formal describing of problems into machine language for subsequent inference. Inverse Parser translates machine language back into formal language, facilitating the verification and checking of the solution process.

\textbf{Data} preserves all details of the problem-solving process and comprises 2 sub-modules. The Problem module ensures the correctness and consistency of the problem input conditions, implementing automatic diagram construction, condition auto-expansion, and validity checks. The Condition module is responsible for data storage.

\textbf{AI Interface} defines the interface for interaction between the AI system and FGPS. Both the AI Automatic Formalization and the AI Problem Solver can be seamlessly integrated with FGPS.

Guided by GFT and modular design, FGPS boasts exceptional extensibility beyond its fundamental features like formal language parsing, GPL execution, human-readable problem-solving processes, and structured output. By invoking FGPS's core modules, we've developed both an interactive solver and a search-based problem solver. Next, we will introduce the forward search algorithm and the backward search algorithm.

\begin{figure}
\centering
\includegraphics[width=\textwidth]{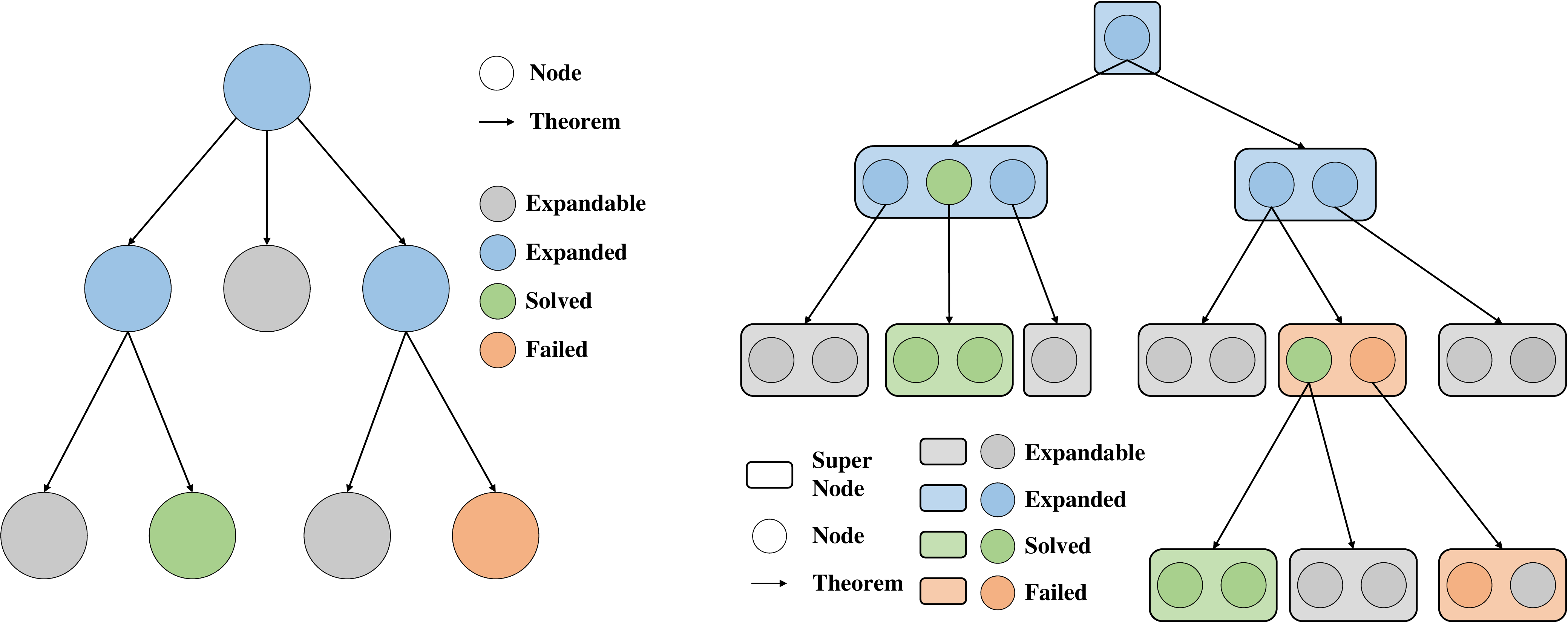}
\caption{Forward search Tree (left). Backward search Tree (right).}
\label{fig-search_tree}
\end{figure}

Forward search starts from the known conditions of the problem and continuously apply theorems to derive new conditions until the goal is achieved. The search process involves the construction of a search tree, with nodes representing sets of known conditions and edges denoting theorems, as depicted in Fig.~\ref{fig-search_tree}. The description of the forward search algorithm is provided in Alg.~\ref{alg-fs}. The function $get\_expandable()$ traverses the search tree based on pre-defined strategies (BFS, DFS, RS and BS) and returns nodes with the EXPANDABLE state. The function $apply\_theorem()$ applies the theorem associated with the current node and returns whether the problem solved. The function $get\_theorem\_seqs()$ returns a list of theorems applied from the root node to the current node. The function $expand()$, guided by the known conditions of the current node, checks the list of applicable theorems and extends new nodes.

\begin{algorithm}
\caption{Forward search}
\label{alg-fs}
\begin{algorithmic}
\STATE \textbf{Input:} $tree$: a tree with the known problem conditions as the root node.
\STATE \textbf{Output:} $theorem\_seqs$: list of theorem sequence for problem solving.
\STATE Initialize an list $theorem\_seqs$
\STATE $node \leftarrow tree.get\_expandable()$
\WHILE{$node$ is not $None$}
    \STATE $solved \leftarrow node.apply\_theorem()$
    \IF{$solved$}
        \STATE $node.state \leftarrow $ SOLVED
        \STATE $theorem\_seqs \leftarrow node.get\_theorem\_seqs()$
        \STATE break
    \ENDIF
    \STATE $node.state \leftarrow $ EXPANDED
    \STATE $l \leftarrow node.expand()$
    \IF{$l = 0$}
        \STATE $node.state \leftarrow $ FAILED
    \ENDIF
    \STATE $node \leftarrow tree.get\_expandable()$
\ENDWHILE
\end{algorithmic}
\end{algorithm}

Backward search (BW), on the other hand, begins with the problem-solving goal, expands it into multiple sub-goals, and repeats this process until all sub-goals are resolved. The search process involves the construction of a search tree, with nodes representing subgoals, hypernodes representing sets of subgoals, and edges representing theorems, as illustrated in Fig.~\ref{fig-search_tree}. The description of the backward search algorithm is provided in Alg.~\ref{alg-bs}. The function $get\_expandable()$ traverses the search tree based on pre-defined strategies, returning hypernodes with the EXPANDABLE state. The function $node.check()$ updates the state of the superNode based on the known problem conditions, while the function $super_node.check()$ updates its own state based on the states of its nodes. The function $expand()$ extends the current goal into several subgoals based on the list of theorems. The function $update()$ propagates the state update from child nodes to parent nodes, starting from the leaves and progressing up to the root. The function $get\_theorem\_seqs()$ provides a list of theorems applied from the current node to the root node.

\begin{algorithm}
\caption{Backward search}
\label{alg-bs}
\begin{algorithmic}
\STATE \textbf{Input:} $super\_tree$:  a super tree with the problem goal as the root super node.
\STATE \textbf{Output:} $theorem\_seqs$: list, theorem sequence for problem solving.
\STATE Initialize an list $theorem\_seqs$
\STATE $super\_node \leftarrow super\_tree.get\_expandable()$
\WHILE{$super\_node$ is not $None$}
    \FOR{$i = 1$ to $len(super\_node.nodes)$}
        \STATE $node \leftarrow super\_node.nodes[i]$
        \STATE $node.check()$
        \IF{$node.state$ is SOLVED}
            \STATE continue
        \ELSIF{$node.state$ is FAILED}
            \STATE break
        \ELSE
            \STATE $l \leftarrow node.expand()$
            \IF{$l = 0$}
                \STATE $node.state \leftarrow $ FAILED
                \STATE break
            \ELSE
                \STATE $node.state \leftarrow $ EXPANDED
            \ENDIF
        \ENDIF
    \ENDFOR
    \STATE $super\_node.check()$
    \STATE $super\_tree.update()$
    \IF{$super\_tree.solved$}
        \STATE $theorem\_seqs \leftarrow super\_tree.get\_theorem\_seqs()$
        \STATE break
    \ENDIF
    \STATE $super\_node \leftarrow super\_tree.get\_expandable()$
\ENDWHILE
\end{algorithmic}
\end{algorithm}

We conducted experiments on the formalgeo7k. Search time and search step of different search methods and strategies can be found in Tab.~\ref{tab-results}.

\begin{table}
\centering
\caption{Experimental Results}
\label{tab-results}
\begin{tabular}{ccccccccccc}
\hline
\multirow{2}{*}{metric} & \multirow{2}{*}{group} & \multirow{2}{*}{method} & \multirow{2}{*}{strategy} & \multicolumn{7}{c}{data} \\
\cline{5-11}
& & & & total & $l_1$ & $l_2$ & $l_3$ & $l_4$ & $l_5$ & $l_6$ \\
\hline
\multirow{16}{*}{time} & \multirow{8}{*}{solved} & FW & BFS & 185.05 & 121.97 & 205.51 & 333.33 & 331.26 & 243.92 & 251.11 \\
& & FW & DFS & 132.35 & 84.11 & 158.91 & 254.29 & 196.02 & 230.39 & 218.05 \\
& & FW & RS & 92.21 & 57.38 & 89.55 & 177.34 & 189.74 & 166.96 & 199.31 \\
& & FW & BS & 58.76 & 35.11 & 81.52 & 106.97 & 207.55 & 191.40 & 121.36 \\
& & BW & BFS & 57.67 & 30.00 & 94.97 & 152.26 & 147.55 & 193.21 & 134.65 \\
& & BW & DFS & 46.45 & 26.34 & 81.17 & 115.83 & 94.85 & 133.23 & 0.85 \\
& & BW & RS & 65.82 & 42.79 & 101.71 & 156.27 & 145.50 & 202.37 & 3.37 \\
& & BW & BS & 75.55 & 47.16 & 121.17 & 172.50 & 165.80 & 207.28 & 146.26 \\
\cline{2-11}
& \multirow{8}{*}{unsolved} & FW & BFS & 574.94 & 435.34 & 583.68 & 615.29 & 624.07 & 662.96 & 680.99 \\
& & FW & DFS & 548.81 & 418.77 & 545.90 & 594.45 & 605.73 & 641.54 & 655.50 \\
& & FW & RS & 542.46 & 394.49 & 545.83 & 591.90 & 606.38 & 632.24 & 649.77 \\
& & FW & BS & 355.74 & 277.24 & 345.70 & 358.92 & 430.08 & 446.34 & 451.65 \\
& & BW & BFS & 579.24 & 549.25 & 572.43 & 592.65 & 588.48 & 596.18 & 597.49 \\
& & BW & DFS & 581.50 & 550.13 & 577.12 & 593.69 & 592.71 & 595.34 & 598.39 \\
& & BW & RS & 580.74 & 554.16 & 578.65 & 589.05 & 588.93 & 592.29 & 597.24 \\
& & BW & BS & 513.86 & 339.10 & 508.03 & 566.64 & 574.09 & 585.96 & 578.42 \\
\hline
\multirow{16}{*}{step} & \multirow{8}{*}{solved} & FW & BFS & 58.44 & 21.85 & 68.67 & 119.56 & 144.80 & 161.44 & 822.18 \\
& & FW & DFS & 87.16 & 33.15 & 95.95 & 225.57 & 232.25 & 257.59 & 727.17 \\
& & FW & RS & 41.89 & 19.14 & 47.63 & 64.32 & 108.26 & 143.41 & 611.00 \\
& & FW & BS & 18.64 & 10.99 & 25.97 & 37.86 & 46.19 & 48.44 & 541.00 \\
& & BW & BFS & 15.14 & 20.68 & 4.80 & 5.60 & 2.31 & 1.00 & 1.00 \\
& & BW & DFS & 5.17 & 4.96 & 5.83 & 5.86 & 3.94 & 1.00 & 1.00 \\
& & BW & RS & 4.34 & 5.02 & 3.13 & 2.81 & 1.27 & 1.00 & 1.00 \\
& & BW & BS & 1.67 & 1.49 & 1.94 & 3.08 & 1.25 & 1.00 & 1.00 \\
\cline{2-11}
& \multirow{8}{*}{unsolved} & FW & BFS & 63.65 & 45.66 & 61.73 & 76.33 & 77.90 & 56.90 & 65.42 \\
& & FW & DFS & 154.25 & 89.57 & 148.09 & 154.30 & 152.47 & 205.00 & 376.90 \\
& & FW & RS & 66.38 & 65.22 & 68.86 & 75.37 & 56.18 & 62.72 & 62.72 \\
& & FW & BS & 37.44 & 27.72 & 36.11 & 41.85 & 42.56 & 40.07 & 54.67 \\
& & BW & BFS & 266.77 & 1148.76 & 195.45 & 21.53 & 10.16 & 4.49 & 59.20 \\
& & BW & DFS & 517.90 & 2656.32 & 154.26 & 6.99 & 2.18 & 1.62 & 17.20 \\
& & BW & RS & 181.74 & 830.49 & 113.67 & 15.27 & 2.14 & 1.34 & 10.16 \\
& & BW & BS & 7.17 & 22.11 & 6.84 & 3.03 & 2.48 & 1.73 & 2.01 \\
\hline
\end{tabular}
\end{table}

\end{document}